\documentclass[10pt,journal]{IEEEtran}
\IEEEoverridecommandlockouts % show footnote comments
\usepackage{blkarray}                                      % to support matrix
\usepackage{graphicx}                                      % include pdf figures
\usepackage{amsmath}
\usepackage{amssymb}
\usepackage{amsfonts}
\usepackage{amsthm}
\usepackage[mathcal]{eucal}
\usepackage{mathrsfs}
\usepackage{booktabs}
\usepackage{enumerate}
\usepackage{multirow}
\usepackage[subrefformat=parens,farskip=0pt,justification=centering]{subfig}
\usepackage{color}
\usepackage{cite}                                          % more citations in one bracket
\usepackage{comment}                                       % use comment
\usepackage{soul}                                          % use highlight command \hl{}
\soulregister\cite7
\soulregister\ref7
\soulregister\pageref7
\usepackage{etoolbox}                                      % commands \newtoggle, \toggletrue, \iftoggle
\usepackage{url}
\usepackage{nth}                                           % nth command
\usepackage{bm}                                            % bm command
\usepackage{courier}
\usepackage{balance}
\usepackage{threeparttable}
\usepackage{xcolor,colortbl}
\usepackage{footnote}
\usepackage{listings}
\usepackage{setspace}                                      % setstretch{}
\usepackage[inline]{enumitem}
\usepackage{verbatim}
\usepackage[bookmarks=false]{hyperref}
\hypersetup{
    colorlinks = true,
    citecolor  = blue,
    linkcolor  = blue,
    urlcolor   = blue,
}
\usepackage{tikz}
\usetikzlibrary{patterns,snakes}
\usetikzlibrary{positioning,calc,fit,decorations.pathmorphing,shapes.geometric, shapes.gates.logic.US, calc}
\usetikzlibrary{arrows,arrows.meta,decorations.markings,shapes,shapes.arrows}
\usetikzlibrary{decorations,decorations.pathreplacing}
\usetikzlibrary{backgrounds}
\usepackage{filecontents}                                  % support to pgfplots
\usepackage{pgfplots}
\usepackage{pgfplotstable}
\usepgfplotslibrary{groupplots}
\usepackage{scalefnt}
\pgfplotsset{compat=newest}
\usepackage{caption}
\usepackage{pifont}                                        % ding commend
\usepackage{cleveref}
\Crefformat{figure}{Fig.~#2#1#3}                           % "Fig.", instead of "Figure"
\Crefname{subfigure}{Fig.}{Figs.}
\Crefname{figure}{Fig.}{Figs.}
\Crefformat{table}{TABLE~#2#1#3}                           % "TABLE", instead of "Table"
\crefname{algorithm}{Algorithm}{Algorithms}
\Crefname{algorithm}{Algorithm}{Algorithms}
\usepackage[figuresright]{rotating}
\usepackage{algorithm}
\iftrue
\usepackage{algpseudocode}                                 % new algorithm package
\algrenewcommand\textproc{\texttt}
\makeatletter
\let\OldStatex\Statex
\renewcommand{\Statex}[1][3]{%
  \setlength\@tempdima{\algorithmicindent}%
  \OldStatex\hskip\dimexpr#1\@tempdima\relax
}
\makeatother
\else
\usepackage{algorithmic}
\fi

\definecolor{CUHKorange}{RGB}{244,106,18} %F47012
\definecolor{CUHKblue}{RGB}{0,111,190}    %006FBE
\definecolor{CUHKgreen}{RGB}{0,127,128}   %007F80
\definecolor{CUHKred}{RGB}{228,46,36}     %E42E24
\definecolor{CUHKyellow}{RGB}{198,148,34} %C69422
\definecolor{CUHKdark}{RGB}{114,44,114}   %722C72
\definecolor{CUHKmiddle}{RGB}{144,44,144} %902C90
\definecolor{CUHKlight}{RGB}{167,44,167} 
\definecolor{CUHKpurple}{RGB}{117,15,109}
\definecolor{CUHKgold}{RGB}{221,163,0}
\definecolor{CUHKribbon}{RGB}{244,223,176}
\definecolor{CUHKblack}{RGB}{34,24,21}

\newcommand{\tool}[1]{$\mathsf{#1}$}
\renewcommand{\bf}[1]{\textbf{#1}}
\renewcommand{\tt}[1]{\texttt{#1}}
\newcommand{\minisection}[1]{\vspace{.02in}\noindent{\textbf{#1}}.}

\usepackage{tcolorbox}
\tcbuselibrary{skins,breakable}
    {\endtcolorbox}
    {\endtcolorbox}

\newtheorem{myproblem}{\textbf{Problem}}

\newtheorem{mytheorem}{\textbf{Theorem}}
\newtheorem{mycorollary}{\textbf{Corollary}}

\crefname{mytheorem}{Theorem}{Theorems}
\crefname{mylemma}{Lemma}{Lemmas}
\crefname{myclaim}{Claim}{Claims}
\crefname{myproperty}{Property}{Properties}
\crefname{mycorollary}{Corollary}{Corollaries}

\RequirePackage[normalem]{ulem} %DIF PREAMBLE
\RequirePackage{color}\definecolor{RED}{rgb}{1,0,0}\definecolor{BLUE}{rgb}{0,0,1} %DIF PREAMBLE
\usepackage{amsmath,amsfonts,bm}

\def\eqref#1{equation~\ref{#1}}
\def\1{\bm{1}}

\def\vd{{\bm{d}}}

\def\vn{{\bm{n}}}

\def\vx{{\bm{x}}}
\def\vy{{\bm{y}}}

\def\vT{{\bm{T}}}

\def\mC{{\bm{C}}}

\def\mK{{\bm{K}}}

\def\mSigma{{\bm{\Sigma}}}

\DeclareMathAlphabet{\mathsfit}{\encodingdefault}{\sfdefault}{m}{sl}
\SetMathAlphabet{\mathsfit}{bold}{\encodingdefault}{\sfdefault}{bx}{n}

\DeclareMathOperator{\diag}{diag}

\graphicspath{{./figs/}{../}}

\usepackage{tikz}
\usepackage{circuitikz}

\begin{document}

\title{
    \tool{FasterRW}: An Efficient Prior-Assisted Random Walk Solver for Thermal Analysis
}

\author{
    Zixiao Wang\textsuperscript{$\ddagger$},
    Tianshu Hou\textsuperscript{$\dagger$},
    Chenghan Wang,
    Zhen Zhuang,
    Leilei Jin,\\
    Tsung-Yi Ho,
    Farzan Farnia,
    Bei Yu\textsuperscript{$\dagger$}\thanks{\textsuperscript{$\dagger$}Corresponding authors. \textsuperscript{$\ddagger$}Zixiao Wang is with The Hong Kong University of Science and Technology (Guangzhou). The other authors are with The Chinese University of Hong Kong.}
}

\maketitle
\thispagestyle{plain}
\pagestyle{plain}
% \linenumbers

\begin{abstract}
Thermal simulation is becoming increasingly critical in modern IC design and manufacturing. Feynman--Kac random walk methods can estimate localized temperatures without solving the global field, but in practical settings without Dirichlet boundaries they require very long paths to control the truncation residual. This paper presents \tool{FasterRW}, a continuation-value framework for accelerating Feynman--Kac thermal random walks. We prove a stopping-time residual identity showing that the expected tail after truncation equals the current attenuation weight multiplied by the temperature at the stopping location. This identity enables a prior-corrected single-point estimator: a cheap, noisy prior replaces the unknown continuation value, making the truncation bias scale with prior error rather than absolute temperature and yielding an error-controlled threshold-selection rule. For multi-point queries, \tool{FasterRW} further converts cross-query pass-through tails into pseudo-samples and refines the resulting observations in prior-error coordinates through kriging. Compared with prior state-of-the-art methods, \tool{FasterRW} achieves up to $28\times$ acceleration while maintaining accuracy. Our code is open-sourced at \url{https://github.com/ShiningSord/FastRW}.
\end{abstract}


\section{Introduction}

Heterogeneous Integration (HI) technologies, such as 3D Integrated Circuits (3D-ICs) in semiconductor packaging, are essential to meet the demands of advanced microelectronics, enabling continued system-level scaling \cite{srcmapt}. However, higher power density in advanced ICs intensifies thermal issues, leading to performance degradation and reduced lifetime \cite{puttaswamy2006thermal}. The heat conduction process, governed by computationally expensive partial differential equations (PDEs), poses significant challenges for thermal management, as exploring the design space may require thousands of simulations to identify optimal parameters \cite{sultan2023varsim, TAP-2.5D, 2025_TCAD_Thermal_placement}.

To mitigate this problem, analytical methods, often leveraging Green's function, have been proposed to provide closed-form expressions for fast temperature estimation on simple geometries \cite{2007_TCAD_Green_Function_1, 2007_TCAD_Green_Function_2}. Numerical methods, such as finite element methods (FEMs), finite difference methods (FDMs), and finite volume methods (FVMs), offer accurate thermal simulation at the cost of fine-grained spatial discretization \cite{MTA_1, MTA_2, ADI_1, ADI_2}. Compact thermal models (CTMs) provide an efficient alternative by representing the thermal behavior based on equivalent RC networks and thereby transforming the problem into SPICE-like simulations, alleviating computational overhead \cite{Huang2006:TVLSI, 3D-ICE, PACT}.

Recently, learning-based thermal predictors have also been explored for fast thermal prediction, including convolutional encoder-decoder models, graph-based predictors, operator-learning methods, and package-level neural simulators \cite{Chhabria2021:ASPDAC, Chen2022:ASPDAC, liu2023deepoheat, Zhang2025:DATE, Hwang2025:TCAD, Wang2025:ICCAD}. After training, these models can provide rapid inference and are attractive for repeated thermal estimation. However, their accuracy and reliability depend heavily on the coverage of the training data, and their behavior under unseen power maps, package geometries, material stacks, or boundary conditions is often harder to certify. In contrast, analytical, numerical, and stochastic solvers are directly derived from the governing PDEs and boundary conditions, which remains important for accuracy-critical thermal analysis.

\begin{figure}[tb!]
    \centering
    \includegraphics[width=.88\linewidth]{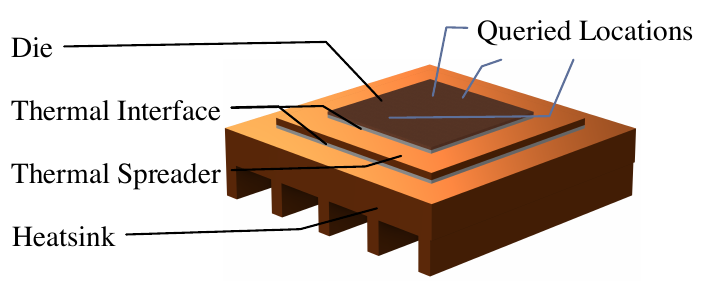}
    \caption{An illustration of a pyramid-shaped IC geometry. In certain scenarios, only the temperatures at a few queried locations are needed.}
    \label{fig:finger}
\end{figure}

Nevertheless, high-resolution temperature data across the entire chip is not always necessary for thermal design, as localized hotspot estimation is typically more critical \cite{zhan2006electrothermal} (see \Cref{fig:finger}).
The stochastic random walk method excels at obtaining localized solutions through parallelized computation and has shown success in thermal analysis and other EDA applications involving Dirichlet boundary conditions (BCs) \cite{Wong2006:DATE,Qian2005:TCAD,Marios2023:TCAD}. The study in \cite{Liang2014:ASPDAC} introduces a hybrid random walk approach capable of handling both adiabatic (Neumann) and convective (Robin) BCs for modeling realistic thermal dissipation. This method was later enhanced in \cite{Yang2022:ICCAD,luo2025gpu,Dong2025:TCAD,Wang2026:DATE} by incorporating the Feynman--Kac formula to improve accuracy. Unlike heuristic treatments, Feynman--Kac-based methods compute the boundary contribution by counting the local time spent near Neumann and Robin boundaries. While effective, these methods require a large number of steps to converge in practical scenarios without Dirichlet boundaries, significantly limiting their efficiency \cite{sawhney2023walk}.

To overcome this limitation, we revisit finite-step Feynman--Kac random walks from the viewpoint of continuation-value estimation. We show that, at any stopping time before absorption, the expected remaining tail is exactly the current Feynman--Kac attenuation weight multiplied by the temperature at the stopping location. This identity changes the role of truncation: instead of choosing an empirical cutoff and accepting an uncontrolled residual, one can estimate the continuation value at the stopped location and make the induced bias explicit.

Building on this observation, we propose \tool{FasterRW}, an efficient random-walk solver for localized thermal analysis. For a single query, \tool{FasterRW} replaces the unknown continuation value with a cheap prior temperature estimate, so the truncation bias is governed by the prior error rather than by the absolute temperature magnitude. This yields an error-controlled threshold-selection rule and permits substantially shorter walks. For multiple queried locations, the same Markov structure exposes additional information: when a path launched from one query passes through another query voxel, its remaining tail can be reused as a pseudo-sample for the visited query. \tool{FasterRW} further refines these direct and reused observations in prior-error coordinates, where the residual of a coarse prior is smoother than the temperature field itself. Our main contributions are summarized as follows:
\begin{itemize}
    \item A stopping-time residual identity for Feynman--Kac-based thermal estimation, showing that the expected truncation tail equals a weighted continuation value at the stopping location.
    \item A prior-corrected single-point estimator with an explicit bias bound and a threshold-selection rule that links the allowable truncation threshold to the prior accuracy.
    \item A multi-point estimation framework that converts cross-query pass-through tails into pseudo-samples and fuses the resulting observations in prior-error coordinates through kriging refinement.
    \item Comprehensive experiments on commonly used 3D thermal benchmarks validating the residual identity, the prior-accuracy/threshold trade-off, the multi-query acceleration, and the end-to-end efficiency of \tool{FasterRW}.
\end{itemize}

\section{Preliminaries}
\label{sec:pre}

In this section, we present the mathematical and algorithmic background necessary for our method. We begin by formulating the steady-state 3D thermal conduction problem with mixed boundary conditions. We then review the application of the Feynman--Kac formula to such problems, which connects the deterministic PDE formulation to stochastic process representations. Finally, we summarize two classes of random-walk-based numerical methods, Walk-on-Grids (WOG) and Walk-on-Spheres (WOS), and their respective roles in simulating heat transfer in source and source-free regions.

\subsection{Problem Formulation}

We consider the steady-state 3D thermal conduction problem governed by the Poisson equation with mixed boundary conditions as in \Cref{eq:problem},
where $T(\vx)$ is the temperature field over the spatial domain $\Omega \subset \mathbb{R}^3$ and $\nabla^2$ denotes the Laplace operator.
\begin{figure}[h]
\begin{equation}
    \begin{aligned}
        -\frac{1}{2} \nabla^2 T(\vx) &= f(\vx), && \vx \in \Omega, \\
        T(\vx) &= \phi_D(\vx), && \vx \in \partial \Omega_D, \\
        \frac{\partial T}{\partial \vn} &= \phi_N(\vx), && \vx \in \partial \Omega_N, \\
        \frac{\partial T}{\partial \vn} + c(\vx) T(\vx) &= \phi_R(\vx), && \vx \in \partial \Omega_R,
        \label{eq:problem}
    \end{aligned}
\end{equation}
\end{figure}
The source term is defined as $f(\vx) = \frac{p(\vx)}{2 k(\vx)}$, with $p(\vx)$ the volumetric heat generation density and $k(\vx)$ the thermal conductivity. The boundary $\partial \Omega$ is decomposed into three disjoint subsets: $\partial \Omega_D$ (Dirichlet), $\partial \Omega_N$ (Neumann), and $\partial \Omega_R$ (Robin), with $\vn$ denoting the outward unit normal vector. The boundary conditions are specified as follows:
\begin{itemize}
    \item \textbf{Dirichlet:} $\phi_D(\vx) = T_D(\vx)$, where $T_D$ is the prescribed temperature.
    \item \textbf{Neumann:} $\phi_N(\vx) = -q_n(\vx)/k(\vx)$, where $q_n(\vx)$ is the prescribed normal heat flux density. An adiabatic boundary corresponds to $q_n(\vx) = 0$.
    \item \textbf{Robin:} $\phi_R(\vx) = c(\vx) T_R(\vx)$ with $c(\vx) = h(\vx)/k(\vx)$, where $h(\vx)$ is the convection coefficient and $T_R(\vx)$ is the reference temperature at the Robin boundary.
\end{itemize}

The domain $\Omega$ may contain multiple materials with different conductivities $k(\vx)$; across material interfaces, both temperature and normal heat flux are continuous. We further distinguish between heat-source regions ($p(\vx) \neq 0$) and source-free regions ($p(\vx) = 0$). With the above notation, the problem of 3D thermal estimation can be formulated as follows.
\begin{myproblem}
Given the Poisson \Cref{eq:problem} with mixed boundary conditions and a set of query locations $\{\vx_q\}$ in $\Omega$, estimate the temperature $T(\vx_q)$ at each query location.
\end{myproblem}

\subsection{Feynman--Kac Method}

The Feynman--Kac formula establishes a connection between certain partial differential equations (PDEs) and expectations over stochastic processes~\cite{oksendal2013stochastic}. In recent years, it has been applied to thermal simulation problems~\cite{Yang2022:ICCAD,Dong2025:TCAD,luo2025gpu}, enabling the solution of the Poisson equation in~\Cref{eq:problem} through Monte Carlo simulation of stochastic particle paths.

Let $X_t$ denote a standard reflecting Brownian motion (SRBM) in the domain $\Omega$, which is reflected at Neumann and Robin boundaries and absorbed at Dirichlet boundaries. The Feynman--Kac representation of the temperature field is
\begin{equation}
\begin{aligned}
    T(\vx) &= \mathbb{E} \left\{ \int_0^\infty \hat{e}_c(t) f(X_t) \, \mathrm{d}t \right\} \\
           &\quad + \mathbb{E} \left\{ \hat{e}_c(t_D) \, \phi_D(X_{t_D}) \right\} \\
           &\quad + \mathbb{E} \left\{ \int_0^\infty \hat{e}_c(t) \, \phi_N(X_t) \, \mathrm{d}L(t) \right\} \\
           &\quad + \mathbb{E} \left\{ \int_0^\infty \hat{e}_c(t) \, \phi_R(X_t) \, \mathrm{d}L(t) \right\},
    \label{eq:FK-formula}
\end{aligned}
\end{equation}
where
\begin{equation}
    \hat{e}_c(t) = \exp \left( -\int_0^t c(X_s) \, \mathrm{d}L(s) \right),
\end{equation}
$\mathbb{E}\{\cdot\}$ denotes the expectation over all SRBM paths, $L(t)$ is the boundary local time (which increases when the process is reflected at Neumann or Robin boundaries)~\cite{zhou2015computation,Dong2025:TCAD}, and $t_D$ is the first hitting time of the Dirichlet boundary. We use the convention
\begin{equation}
    L(t)=\lim_{\rho\to 0}\frac{1}{2\rho}\int_0^t \mathbf{1}_{D_\rho}(X_s)\,\mathrm{d}s,
\end{equation}
where $D_\rho$ denotes the $\rho$-width neighborhood of the reflecting boundary inside $\Omega$; this convention absorbs the usual factor $1/2$ in the boundary local-time integrals. The coefficient $c(\vx)$ is given by $c(\vx) = h(\vx)/k(\vx)$ on the Robin boundary and is zero elsewhere. In implementation, the growth of $L(t)$ is evaluated through the random walk behavior inside a thin boundary neighborhood, following the boundary-strip allocation in prior PIRW implementations~\cite{Yang2022:ICCAD,Dong2025:TCAD,luo2025gpu}.

In practice, the infinite integrals in~\Cref{eq:FK-formula} are truncated. Simulation paths terminate in one of two cases:
\begin{enumerate}
    \item the path hits the Dirichlet boundary, or
    \item the Feynman--Kac functional $\hat{e}_c(t)$ decreases below a pre-defined threshold as Robin-boundary local time accumulates.
\end{enumerate}
For problems where Dirichlet boundaries are present, paths usually terminate at a Dirichlet boundary before the functional $\hat{e}_c(t)$ becomes small enough. For problems without Dirichlet boundary conditions, which are common in practice, Feynman--Kac-based methods achieve higher accuracy than heuristic approaches~\cite{Liang2014:ASPDAC}.
However, in such cases, a large number of simulation steps may be required to sufficiently reduce $\hat{e}_c(t)$ and control the truncation error, resulting in high computational cost. Furthermore, the choice of path length heavily depends on empirical experience, and \textbf{there is currently no clear guideline on when a path should be terminated}. One of the key contributions of this work is to provide a quantitative analysis of the truncation error, together with a novel approach that reduces the number of required simulation steps and paths while preserving accuracy.

\begin{figure}[tb!]
    \centering
    \includegraphics[width=.818\linewidth]{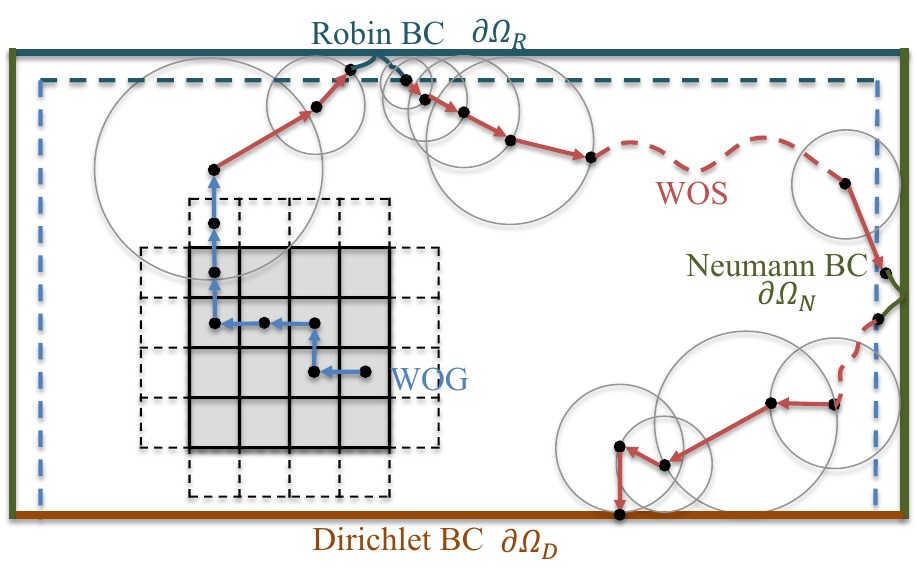}
    \caption{An illustration of the random walk path in \tool{FasterRW}.}
    \label{fig:rw}
\end{figure}

\subsection{Random Walk Methods for Thermal Simulation}

In the context of thermal simulation, the design of random-walk paths can be broadly categorized into two classes: \emph{Walk-on-Grids} and \emph{Walk-on-Spheres}. 

\minisection{Walk-on-Grids}
In WOG methods, the heat-source region is discretized into a uniform voxel grid, with each voxel assigned constant temperature and power density values. Brownian motion is approximated by random jumps between adjacent grid nodes. For two neighboring nodes $a$ and $b$, the probability of transition from $a$ to $b$, denoted $P(a,b)$, is given by
\begin{equation}
    P(a,b) = \frac{k(a,b)}{\sum_{x \in \mathrm{adj}(a)} k(a,x)}, 
    \quad 
    k(a,b) = \frac{2 h_x h_y}{\frac{h_{za}}{k_a} + \frac{h_{zb}}{k_b}},
\end{equation}
where $\mathrm{adj}(a)$ denotes the set of nodes adjacent to $a$; $k_a$ and $k_b$ are the thermal conductivities at $a$ and $b$, respectively; $h_x$ and $h_y$ are the side lengths of the shared interface; and $h_{za}$ and $h_{zb}$ are the edge lengths (thicknesses) of the two voxels along the direction normal to the shared interface (i.e., perpendicular to that interface).

\minisection{Walk-on-Spheres}
WOS methods are particularly efficient in homogeneous, source-free regions. In such regions, the temperature at a point $\vx$ equals the average temperature over any sphere centered at $\vx$:
\begin{equation}
    T(\vx) = \frac{1}{4\pi} \int_0^{2\pi} \int_0^\pi 
    T(r,\omega,\theta) \, \sin \theta \, \mathrm{d}\theta \, \mathrm{d}\omega,
\end{equation}
where $r$ is the sphere radius, $\omega$ and $\theta$ are the azimuthal and polar angles, respectively, and $T(r,\omega,\theta)$ is the boundary temperature on the spherical surface. Each WOS step moves from the current center $\vx$ to a uniformly sampled point $\vy$ on the sphere surface. The next step repeats the process with $\vy$ as the new center. For efficiency, the radius $r$ is chosen as large as possible without crossing a boundary.

In this work, we adopt WOG for simulating heat transfer in the heat-source region and WOS for the source-free region, following established practice in the literature~\cite{Liang2014:ASPDAC,Yang2022:ICCAD}. An illustration of the random walk process can be found in \Cref{fig:rw}.

\section{FasterRW Method}
\label{sec:method}

\begin{figure*}[t]
    \centering
    \includegraphics[width=.7\linewidth]{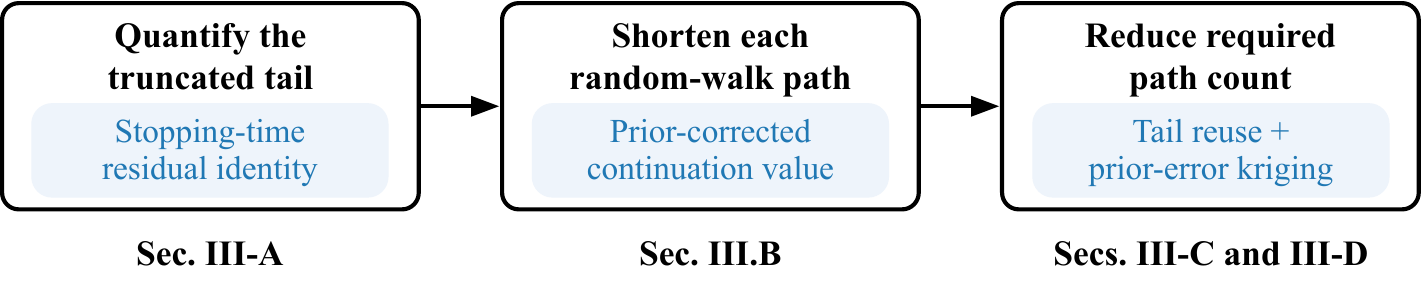}
    \caption{\tool{FasterRW} method overview.} \label{fig:overview}
\end{figure*}

The method is organized around one question: after a reflected Feynman--Kac random-walk path is stopped early, what information is needed to replace its remaining tail? As illustrated in \Cref{fig:overview}, we first show that the expected tail has an exact continuation-value form at any stopping time in \Cref{subsec:truncation}. This residual identity then leads to three estimators: a prior-corrected single-point estimator that controls truncation bias (\Cref{subsec:single}), a tail-reuse pseudo-sample for multi-point queries (\Cref{subsec:multi}), and a prior-error-space kriging refinement that exploits spatial coherence across queried points (\Cref{subsec:prior_error_coordinates}).

\subsection{Truncation as Continuation-Value Estimation}
\label{subsec:truncation}

This subsection establishes the mathematical basis of \tool{FasterRW}. Instead of treating early stopping as an empirical approximation, we show that the expected remaining Feynman--Kac tail at any stopping time is exactly the attenuation weight times the steady-state temperature at the stopping location. This converts path truncation into a continuation-value estimation problem: controlling the truncation error amounts to estimating the temperature at the random stopping location accurately enough.

As discussed in \Cref{sec:pre}, problems without Dirichlet boundaries are the most common in practice, and are precisely the cases where Feynman--Kac-based methods demonstrate significant accuracy advantages. We focus on the reflecting-dominated regime where the SRBM path does not encounter a Dirichlet boundary before the Feynman--Kac functional $\hat{e}_c(t)$ becomes sufficiently small. If a Dirichlet boundary is reached earlier, the path terminates by absorption and no truncation error is incurred.

In the Feynman--Kac representation (\Cref{eq:FK-formula}), for a given positive threshold $\Lambda \in (0,1)$, we define $t_\Lambda$ as the first time at which $\hat{e}_c(t) \leq \Lambda$. The single-path contribution can then be decomposed into two parts:
\begin{equation}
    T_i(\vx) = \mathbb{P}_0^{t_\Lambda} + \mathbb{P}_{t_\Lambda}^\infty,
    \label{eq:twopart}
\end{equation}
where $T_i(\vx)$ denotes one realization of the temperature at the starting point $\vx$, $\mathbb{P}_0^{t_\Lambda}$ aggregates the source and boundary contributions collected before $t_\Lambda$, and $\mathbb{P}_{t_\Lambda}^\infty$ represents the remaining contribution after $t_\Lambda$. If the path is truncated at $t_\Lambda$, the resulting truncation error is exactly $\mathbb{P}_{t_\Lambda}^\infty$.

To quantify this error, we formalize the following residual identity.
\begin{mytheorem}[Stopping-time truncation residual identity]
\label{thm:truncation_residual}
Let $\{X_t\}_{t\ge 0}$ be the reflected process defined in \Cref{sec:pre} with natural filtration $\{\mathcal{F}_t\}_{t\ge 0}$, and let $\tau$ be a stopping time such that $\tau<t_D$ on the event of interest. Then the truncation tail satisfies
\begin{equation}
    \mathbb{E}\!\left[\mathbb{P}_{\tau}^{\infty}\mid \mathcal{F}_{\tau}\right]
    =
    \hat{e}_c(\tau)\,T\!\left(X_{\tau}\right).
    \label{eq:error}
\end{equation}
\end{mytheorem}

This identity is the central object used by \tool{FasterRW}. It says that \textbf{the unsimulated tail is not an arbitrary leftover term: in expectation, it is a continuation value at the stopping location, scaled by the current attenuation weight.}

\begin{mycorollary}
\label{corollary:error}
Given the truncation location $X_t$ and the corresponding functional $\hat{e}_c(t)$, the expected truncation error equals the weighted temperature $\hat{e}_c(t)\,T(X_t)$ at the truncation location. In particular, for the threshold rule $\tau=t_\Lambda$,
\begin{equation}
    \mathbb{E}\!\left[\mathbb{P}_{t_\Lambda}^{\infty}\mid \mathcal{F}_{t_\Lambda}\right]
    =
    \hat{e}_c(t_\Lambda)\,T(X_{t_\Lambda})
    \le
    \Lambda\,T(X_{t_\Lambda}).
\end{equation}
\end{mycorollary}

The proof follows from the strong Markov property of the reflected process and is provided in \Cref{app:truncation_proof}. An intuitive way to interpret \Cref{eq:error} is that the truncation tail behaves like the contribution induced by imposing a virtual Dirichlet boundary condition at the truncation position $X_{t_\Lambda}$. Unlike previous works, which select $\Lambda$ empirically, this identity provides a principled guideline: to maintain the truncation error within an acceptable range, the threshold should be chosen such that the expected residual scale, $\Lambda\,\mathbb{E}\!\left[T(X_{t_\Lambda})\right]$, remains sufficiently small.

However, in problems with only Neumann or Robin boundaries, $\hat{e}_c(t)$ decreases only upon reflections, meaning that a large number of SRBM steps are required before termination. For example, in one of our test cases, reducing $\hat{e}_c(t)$ to $10\%$ of its initial value required roughly $1.5\times 10^6$ random walk steps. The next subsection uses the continuation-value view to stop such paths earlier while keeping the induced bias explicit.

\subsection{Prior-Corrected Single-Point Estimation}
\label{subsec:single}

This subsection uses the residual identity to shorten a single random-walk path. The exact continuation value is not available during simulation, so we replace it with a cheap prior estimate at the stopping location. The key effect is to make the remaining bias depend on the prior's error, rather than on the much larger absolute temperature scale.

We assume that a \emph{prior} temperature estimate $\tilde{T}(\vx)$ is available within a subdomain $\Omega_s \subseteq \Omega$ before initiating the random walk process:
\begin{equation}
    \tilde{T}(\vx) = T(\vx) + \epsilon(\vx), \quad \vx \in \Omega_s,
    \label{eq:prior_error_convention}
\end{equation}
where $\epsilon(\vx)$ denotes the prior estimation error. The procedure for obtaining such prior estimates will be described in a later section.

With this prior estimation in place, the random walk path in \tool{FasterRW} terminates under one of the following two conditions:
\begin{enumerate}
    \item The path reaches a Dirichlet boundary, or
    \item The Feynman--Kac functional $\hat{e}_c(t)$ decreases below a predefined threshold as the local time $L(t)$ increases, \emph{and} the current location lies in the subdomain $\Omega_s$ where the prior temperature field is available.
\end{enumerate}

\begin{figure}[t!]
    \centering
    \includegraphics[width=.68\linewidth]{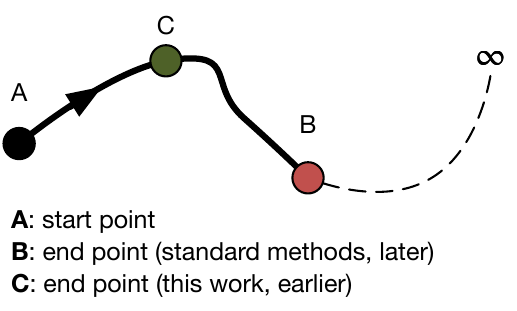}
    \caption{A random walk path in a single-point simulation. Standard methods compute the integration from A to B. In contrast, \tool{FasterRW} computes the integration from A to C and estimates the remaining contribution from C to infinity.}

    \label{fig:single}
\end{figure}

When termination occurs under the second condition at a stopping time $\tau$, \tool{FasterRW} does not simply return the accumulated term $\mathbb{P}_0^\tau$ as in previous approaches. Instead, it augments this value using the prior estimation, yielding the \emph{prior-corrected single-path estimator}
\begin{equation}
    \begin{aligned}
        \hat{T}_i(\vx) &\;\triangleq\; \mathbb{P}_0^\tau + \hat{e}_c(\tau)\,\tilde{T}(X_\tau) \\
        &\;=\; \mathbb{P}_0^\tau + \mathbb{E}[\mathbb{P}_\tau^\infty \mid \mathcal{F}_\tau] + \hat{e}_c(\tau)\,\epsilon(X_\tau),
    \end{aligned}
    \label{eq:single}
\end{equation}
where $\hat{T}_i(\vx)$ is the $i$-th single-path estimate of the temperature at the query location $\vx$, $\mathbb{P}_0^\tau$ is the Feynman--Kac sum accumulated up to the truncation time $\tau$, $\tilde{T}(X_\tau)$ is the prior estimate at the truncation location $X_\tau$, and $\epsilon(X_\tau)$ is the prior-estimation error there (cf.~\Cref{eq:prior_error_convention}). Compared with vanilla truncation, the expected truncation error is reduced from $\hat{e}_c(\tau)\,T(X_\tau)$ to $\hat{e}_c(\tau)\,\epsilon(X_\tau)$. Since in most cases $|\epsilon(X_\tau)| \ll T(X_\tau)$, \textbf{the allowable threshold $\Lambda$ in \tool{FasterRW} can be chosen much larger than in conventional methods, significantly reducing the number of random walk steps required to achieve the same truncation error}, as illustrated in \Cref{fig:single}. The following result makes this bias reduction explicit.

This estimator is the operational rule used by each accelerated path. Instead of discarding the unsimulated tail, it returns the simulated prefix plus a prior-based continuation estimate, which is why a path can stop much earlier than in vanilla Feynman--Kac random walk.

For comparison, define the ideal but infeasible truncation-corrected estimator that replaces the prior with the exact continuation value:
\begin{equation}
    \hat{T}_i^\star(\vx) \;\triangleq\; \mathbb{P}_0^\tau + \hat{e}_c(\tau)\,T(X_\tau),
    \label{eq:ideal_estimator}
\end{equation}
which is the oracle counterpart of $\hat{T}_i(\vx)$ obtained by substituting the true temperature $T(X_\tau)$ for the prior $\tilde{T}(X_\tau)$. By \Cref{thm:truncation_residual}, $\hat{T}_i^\star(\vx)$ is unbiased for $T(\vx)$.

\begin{mytheorem}[Bias induced by prior correction]
\label{thm:plugin_bias}
Let $\hat{T}_i(\vx)$ and $\hat{T}_i^\star(\vx)$ be defined in \Cref{eq:single,eq:ideal_estimator}. Then
\begin{equation}
    \hat{T}_i(\vx) - \hat{T}_i^\star(\vx) = \hat{e}_c(\tau)\,\epsilon(X_\tau),
    \label{eq:plugin_error_pathwise}
\end{equation}
and the bias satisfies
\begin{equation}
    \mathbb{E}\!\left[\hat{T}_i(\vx)\right] - T(\vx)
    = \mathbb{E}\!\left[\hat{e}_c(\tau)\,\epsilon(X_\tau)\right].
    \label{eq:plugin_bias}
\end{equation}
\end{mytheorem}

\begin{mycorollary}[Uniform bound under threshold truncation]
\label{cor:plugin_bias_uniform}
Suppose the truncation rule enforces $\hat{e}_c(\tau) \le \Lambda$ almost surely and the prior error satisfies $\sup_{\vx\in\Omega_s}|\epsilon(\vx)| \le \epsilon_{\max}$. Then
\begin{equation}
    \left|\mathbb{E}\!\left[\hat{T}_i(\vx)\right] - T(\vx)\right|
    \le \Lambda\,\epsilon_{\max}.
\end{equation}
Equivalently, to ensure an absolute bias tolerance $b_{\mathrm{tol}}$, it suffices to choose
\begin{equation}
    \Lambda \le \frac{b_{\mathrm{tol}}}{\epsilon_{\max}}.
    \label{eq:lambda_rule}
\end{equation}
\end{mycorollary}

\Cref{cor:plugin_bias_uniform} gives a direct threshold-selection rule: once a prior error bound is available, $\Lambda$ can be chosen from the target truncation-induced bias rather than by empirical tuning. A detailed bias--variance decomposition of this plug-in correction is given in \Cref{app:plugin_bias_variance}.

Once all paths have terminated, the estimated temperature at the starting point, $\bar{T}(\vx)$, is obtained by averaging the individual path estimators $\{\hat{T}_i(\vx)\}_{i=1}^N$ from \Cref{eq:single}:
\begin{equation}
\bar{T}(\vx) = \frac{1}{N} \sum_{i=1}^N \hat{T}_i(\vx),
\label{eq:avg}
\end{equation}
where $\bar{T}(\vx)$ is the empirical mean used as the final temperature estimate at $\vx$, $N$ is the total number of paths initiated from $\vx$, and $\hat{T}_i(\vx)$ is the prior-corrected single-path estimator of the $i$-th path.
Since the prior estimation approach in \Cref{eq:single} applies a larger truncation threshold $\Lambda$ and shortens the path length, the resulting variance of individual paths is also smaller than in conventional methods.

By the law of large numbers, $N$ should be chosen so that the variance of the mean remains within an acceptable range:
\begin{equation}
\mathrm{Var}\!\left[\bar{T}(\vx)\right] = \frac{\sigma^2(\vx)}{N},
\label{eq:largenumber}
\end{equation}
where $\sigma^2(\vx) \triangleq \mathrm{Var}[\hat{T}_i(\vx)]$ is the per-path variance of the prior-corrected single-path estimator at $\vx$. In practice, $N$ is selected to balance computational cost and accuracy, as increasing $N$ reduces variance but requires more computation.

Operationally, the single-point stage leaves the WOG/WOS transition rule unchanged and only modifies the stopping rule and returned path contribution: a path exits at a Dirichlet boundary as usual, or exits once $\hat{e}_c(t)\le\Lambda$ at a location where the prior is available, in which case it returns $\mathbb{P}_0^{\tau}+\hat{e}_c(\tau)\,\tilde{T}(X_\tau)$ instead of the uncorrected partial sum.

\minisection{Obtaining Prior Estimation}
The prior field $\tilde{T}$ only needs to be inexpensive and reasonably accurate over a subdomain $\Omega_s$; it is not required to match the accuracy of the final random-walk estimate. Its role is to reduce the correction scale in \Cref{eq:single} from the absolute temperature to the prior error, so even a coarse prior can be useful. We examine the dependence between prior accuracy and the allowable truncation threshold in \Cref{tab:tradeoff}.

In \tool{FasterRW}, we use a coarse FEM solve as the prior. Such a solve is inexpensive because the element count can be kept low, while still providing enough accuracy to reduce the truncation residual. In our chip thermal experiments, the coarse FEM prior typically has a maximum error around $5\,\mathrm{K}$, compared with an on-chip average temperature around $400\,\mathrm{K}$; this allows a truncation threshold up to about $80\times$ larger than in conventional random walk methods.

As described in \Cref{sec:pre}, we apply WOG in the source region and WOS in the source-free region, using prior predictions only in the source region. When a path terminates at a source-region voxel, the prior temperature at that voxel is returned as specified in \Cref{eq:single}.

\subsection{Tail-Reuse Observations for Multi-Point Queries}
\label{subsec:multi}

\begin{figure}[t!]
    \centering
    \includegraphics[width=0.68\linewidth]{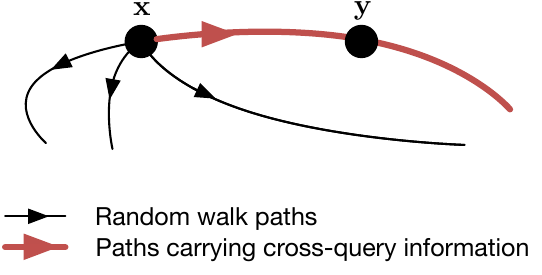}
    \caption{Random walk paths in multi-point simulation. Some paths contain cross-point information and can correct the prediction.}

    \label{fig:multi}
\end{figure}

This subsection extends the continuation-value view from path endpoints to intermediate pass-through events. As illustrated in \Cref{fig:multi}, in a multi-query run, when a path launched from one query voxel enters another query voxel, its remaining segment behaves like a fresh continuation path from the visited location. The already simulated tail can therefore be reused as an additional observation for the visited query, reducing the number of direct paths needed per query.

When the query set contains multiple locations $\{\vx_q\}_{q=1}^M$, conventional approaches estimate each temperature $T(\vx_q)$ using only paths that originate from $\vx_q$ (cf.~\Cref{eq:avg}). The truncation residual identity of \Cref{thm:truncation_residual} reveals that paths starting elsewhere can also carry information about $T(\vx_q)$.

Consider the $n$-th path that starts at query point $\vx_i$, and let $\hat{T}_{i,n}$ denote the prior-corrected single-path estimator from \Cref{eq:single} evaluated along this specific path (the realization of $\hat{T}_i(\vx_i)$ on the $n$-th path from origin $i$). In continuous space, hitting an exact query point has probability zero, so we treat ``visiting $\vx_j$'' as the event that the walk enters the discretization voxel associated with $\vx_j$. Suppose the path enters such a voxel at time $t_k$, where the subscript $k$ indexes successive pass-through events along the path. By the strong Markov property invoked in the proof of \Cref{thm:truncation_residual}, the remainder of the path is, in distribution, a fresh reflected process started from the SRBM position $X_{t_k}$ with its Feynman--Kac weight scaled by $\hat{e}_c(t_k)$. The complete prior-corrected single-path estimator therefore admits the pathwise decomposition
\begin{equation}
    \hat{T}_{i,n} \;=\; \mathbb{P}_0^{t_k} + \mathbb{P}_{t_k}^{\infty,\,n},
    \label{eq:tail_decomp}
\end{equation}
where $\mathbb{P}_{t_k}^{\infty,\,n}$ denotes the post-$t_k$ contribution along this single path (including the prior-assisted truncation correction in \Cref{eq:single}). Solving \Cref{eq:tail_decomp} for $\mathbb{P}_{t_k}^{\infty,\,n}$ and dividing by $\hat{e}_c(t_k)$ produces a pseudo-sample of $T(X_{t_k})$, the temperature at the visited voxel.

\begin{mycorollary}[Tail-reuse pseudo-sample]
\label{cor:tail_reuse}
For each pass-through event $k$ along the $n$-th path starting at $\vx_i$, define the tail-reuse pseudo-sample
\begin{equation}
    S_{j_k,\,k} \;\triangleq\;
    \frac{\mathbb{P}_{t_k}^{\infty,\,n}}{\hat{e}_c(t_k)}
    =
    \frac{\hat{T}_{i,n} - \mathbb{P}_0^{t_k}}{\hat{e}_c(t_k)},
    \label{eq:tail_reuse}
\end{equation}
where $j_k$ is the index of the query voxel entered at $t_k$, $\mathbb{P}_{t_k}^{\infty,\,n}$ is the post-$t_k$ tail contribution from \Cref{eq:tail_decomp}, $\hat{T}_{i,n}$ is the complete prior-corrected single-path estimator, $\mathbb{P}_0^{t_k}$ is the pre-$t_k$ Feynman--Kac sum, and $\hat{e}_c(t_k)$ is the Feynman--Kac weight at $t_k$.

This definition is the conversion rule from a pass-through event to a reusable observation. It removes the contribution accumulated before the path reached the visited query and rescales the remaining tail back to the temperature scale of that query.

Then $S_{j_k,k}$, viewed as an estimate of $T(\vx_{j_k})$, satisfies
\begin{equation}
    \mathbb{E}\!\left[S_{j_k,\,k} \;\middle|\; X_{t_k}=\vy\right]
    \;=\; T(\vy)
    \;+\; \mathbb{E}\!\left[\tfrac{\hat{e}_c(\tau)}{\hat{e}_c(t_k)}\,\epsilon(X_\tau)\right],
    \label{eq:tail_reuse_mean}
\end{equation}
and
\begin{equation}
    \mathrm{Var}\!\left[S_{j_k,\,k} \;\middle|\; X_{t_k}=\vy\right] \;=\; \sigma^2(\vy),
    \label{eq:tail_reuse_var}
\end{equation}
where $\tau$ is the post-$t_k$ truncation time of the path and $\sigma^2(\vy)$ is the per-path variance of the prior-corrected single-path estimator at $\vy$ from \Cref{eq:largenumber}.
\end{mycorollary}

\Cref{cor:tail_reuse} follows from \Cref{thm:truncation_residual,thm:plugin_bias} applied to the post-$t_k$ segment of the path. The pseudo-sample $S_{j_k,k}$ is therefore approximately unbiased for $T(\vx_{j_k})$, with a residual plug-in bias inherited from \Cref{cor:plugin_bias_uniform} and amplified by the factor $1/\hat{e}_c(t_k)$. To control this amplification, we restrict to events with $\hat{e}_c(t_k) \ge e_{\min}$ (default $e_{\min}=0.3$). When the same path enters the same target voxel more than once, only the record with the largest $\hat{e}_c(t_k)$ is retained, so that pseudo-samples within a target are uncorrelated.

\minisection{Per-target Bayesian fusion}
Let $\mathcal{S}_q = \{S_{q,\,k} : j_k = q,\ \hat{e}_c(t_k) \ge e_{\min}\}$ denote the set of tail-reuse pseudo-samples collected at the query point $\vx_q$, with cardinality $n_q \triangleq |\mathcal{S}_q|$. By \Cref{cor:tail_reuse}, each pseudo-sample $S_{q,k}$ has the same conditional variance as a single direct path estimator $\hat{T}_{q,n}$ from $\vx_q$. Modelling both as conditionally Gaussian and placing an improper flat prior on $T(\vx_q)$,
\begin{equation}
    \begin{aligned}
        \hat{T}_{q,n} \mid T(\vx_q) &\sim \mathcal{N}\!\bigl(T(\vx_q),\,\sigma^2(\vx_q)\bigr), \\
        S_{q,k} \mid T(\vx_q) &\sim \mathcal{N}\!\bigl(T(\vx_q),\,\sigma^2(\vx_q)\bigr),
    \end{aligned}
    \label{eq:per_target_obs}
\end{equation}
where $\sigma^2(\vx_q)$ is the per-path variance at $\vx_q$ (cf.~\Cref{eq:largenumber}).

The posterior mean of $T(\vx_q)$ is the inverse-variance weighted combination of the direct mean $\bar{T}(\vx_q) \triangleq N^{-1}\sum_n \hat{T}_{q,n}$ and the tail-reuse mean $\bar{S}_q \triangleq n_q^{-1}\sum_k S_{q,k}$:
\begin{equation}
    \hat{T}_q \;=\; \frac{N\,\bar{T}(\vx_q) + n_q\,\bar{S}_q}{N + n_q},
    \qquad
    \mathrm{Var}\!\bigl[\hat{T}_q\bigr] \;=\; \frac{\sigma^2(\vx_q)}{N + n_q},
    \label{eq:fused_estimator}
\end{equation}
where $\hat{T}_q$ is the per-target fused estimate of $T(\vx_q)$. \textbf{Thus, paths launched for one query are no longer useful only for their own starting point: each qualified pass-through tail can reduce the direct path budget of another queried point.}

In practice, $\sigma^2(\vx_q)$ is replaced by the empirical sample variance $\hat{\sigma}^2(\vx_q)$ of $\{\hat{T}_{q,n}\}_{n=1}^N$, and the corresponding empirical sample variance $\widehat{\mathrm{Var}}(\mathcal{S}_q)$ is used for the tail-reuse term, yielding
\begin{equation}
    \hat{T}_q =
    \frac{N\,\bar{T}(\vx_q) + \frac{\hat{\sigma}^2(\vx_q)}{\widehat{\mathrm{Var}}(\mathcal{S}_q)}\,n_q\,\bar{S}_q}
    {N + \frac{\hat{\sigma}^2(\vx_q)}{\widehat{\mathrm{Var}}(\mathcal{S}_q)}\,n_q},
    \,
    \mathrm{Var}\!\bigl[\hat{T}_q\bigr]
    =
    \frac{\hat{\sigma}^2(\vx_q)}
    {N + \frac{\hat{\sigma}^2(\vx_q)}{\widehat{\mathrm{Var}}(\mathcal{S}_q)}\,n_q}.
    \label{eq:fused_estimator_implementation}
\end{equation}
This is the fusion formula used in implementation. It keeps the same idea as the idealized rule, but estimates the relative reliability of direct paths and reused tails from the sampled data rather than assuming identical variances.

Thus the multi-point procedure consists of three statistical steps: record qualified cross-query pass-through events during the usual prior-assisted walks, convert their remaining path contributions into pseudo-samples through \Cref{eq:tail_reuse}, and combine direct and tail-reuse observations through \Cref{eq:fused_estimator_implementation}. The estimator in \Cref{eq:fused_estimator} provides a direct temperature-space fusion of single-point Monte Carlo estimates and tail-reuse pseudo-samples. The key information source exposed by \Cref{cor:tail_reuse} is that every step of a multi-point walk, and not only its termination, contributes statistical information about the queried temperatures; this reduces the number of paths required to attain the desired precision in \tool{FasterRW}. We next reuse the fused estimates and variances as noisy observations of the prior-error field, enabling spatial refinement across the query set.

\subsection{Prior-Error Kriging Refinement}
\label{subsec:prior_error_coordinates}

This subsection refines the multi-point estimates by changing the statistical target from temperature to prior error. The temperature field may contain sharp hotspot structure, while the error of a coarse FEM prior is typically smoother over nearby query locations. We use this smoother error field to regularize the Monte Carlo estimates across a query batch through universal kriging.

The tail-reuse fusion above estimates the queried temperatures directly. Because \tool{FasterRW} already maintains a coarse prior field $\tilde{T}$ for the single-point plug-in correction (\Cref{eq:single}), the same per-target observations can equivalently be expressed using the prior-error convention in \Cref{eq:prior_error_convention}. We model the stacked prior-error vector
\[
    \boldsymbol{\epsilon}\triangleq[\epsilon(\vx_1),\ldots,\epsilon(\vx_M)]^\top
\]
where $\epsilon(\vx_i)$ denotes the prior error at the $i$-th query point. \textbf{Estimating the smoother prior-error field allows \tool{FasterRW} to borrow correction information across nearby queries, further reducing the path count beyond tail reuse alone.}

\minisection{Combined per-target observation}
Let $\hat{T}_q$ be the per-target fused estimate of $T(\vx_q)$ from \Cref{eq:fused_estimator_implementation}. Define the per-target prior-error observation
\begin{equation}
    \begin{aligned}
        d_q &\;\triangleq\; \tilde{T}(\vx_q) - \hat{T}_q, \\
        d_q \mid \epsilon(\vx_q)
        &\sim \mathcal{N}\!\left(\epsilon(\vx_q),\,\mathrm{Var}\!\bigl[\hat{T}_q\bigr]\right),
    \end{aligned}
    \label{eq:prior_error_obs}
\end{equation}
where $d_q$ is the observed discrepancy between the prior $\tilde{T}(\vx_q)$ and the fused estimate $\hat{T}_q$, and $\epsilon(\vx_q)$ is the (unknown) prior-estimation error at $\vx_q$. Stacked over the query set, this gives the vector observation $\vd \triangleq [d_1,\ldots,d_M]^\top$ with conditional distribution $\vd\mid\boldsymbol{\epsilon}\sim\mathcal{N}(\boldsymbol{\epsilon},\,\mSigma_{\mathrm{MC}})$, where $\mSigma_{\mathrm{MC}} \triangleq \diag(\mathrm{Var}[\hat{T}_1],\ldots,\mathrm{Var}[\hat{T}_M])$ is the diagonal Monte Carlo noise covariance.

\minisection{Universal kriging prior}
We model the prior-error vector $\boldsymbol{\epsilon}$ as the sum of an unknown constant trend $\mu$ and a zero-mean spatial Gaussian process:
\begin{equation}
    \boldsymbol{\epsilon} \;=\; \mu\,\mathbf{1} + \mathbf{g},
    \qquad
    \mathbf{g}\sim\mathcal{N}(\mathbf{0},\,\mK_\ell),
    \qquad
    p(\mu)\propto 1,
    \label{eq:uk_prior}
\end{equation}
where $\mu$ is the global drift of the prior error, $\mathbf{1}\in\mathbb{R}^M$ is the all-ones vector, $\mathbf{g}$ is the zero-mean spatial residual with covariance matrix $\mK_\ell$, and $p(\mu)\propto 1$ is an improper flat prior over $\mu$.

The covariance $\mK_\ell$ uses an RBF kernel,
\begin{equation}
    [\mK_\ell]_{ij}
    \;=\;
    \sigma_\epsilon^2 \exp\!\left(
    -\frac{\|\vx_i-\vx_j\|^2}{2\ell^2}
    \right),
    \label{eq:uk_rbf}
\end{equation}
with marginal variance $\sigma_\epsilon^2$ and length scale $\ell$. The trend $\mu$ absorbs any near-uniform bias in the coarse prior, e.g., a global temperature shift between a low-DoF FEM solution and the truth, while $\mathbf{g}$ captures the residual spatial structure of the error. Assigning $\mu$ an improper flat prior lets it be inferred from data without anchoring the posterior toward an arbitrary value.

\minisection{REML hyperparameter selection}
Let $\mC \triangleq \mK_\ell + \mSigma_{\mathrm{MC}}$ denote the total covariance of $\vd$ given $\mu$. Integrating $\mu$ out against its flat prior yields the restricted maximum-likelihood (REML) marginal log-likelihood, used to select the kernel hyperparameters $(\ell,\sigma_\epsilon^2)$:
\begin{align}
\log p(\vd \mid \ell, \sigma_\epsilon^2)
=& -\tfrac{M-1}{2}\log 2\pi - \tfrac{1}{2}\log\det\mC - \tfrac{1}{2}\log A \notag\\
&- \tfrac{1}{2}\,(\vd - \hat{\mu}\mathbf{1})^\top\mC^{-1}(\vd - \hat{\mu}\mathbf{1}).
\label{eq:reml}
\end{align}
Here, the auxiliary quantities are
\begin{equation}
    A \;\triangleq\; \mathbf{1}^\top\mC^{-1}\mathbf{1},
    \qquad
    \hat{\mu} \;\triangleq\; \frac{\mathbf{1}^\top\mC^{-1}\vd}{A},
    \label{eq:c_hat}
\end{equation}
with $A$ the precision of $\hat{\mu}$ under known $\mC$ and $\hat{\mu}$ the generalized-least-squares estimate of the global drift. We maximize \Cref{eq:reml} over a grid in $(\ell,\sigma_\epsilon^2)$, with $\ell$ scaled to the spatial discretization and $\sigma_\epsilon^2$ scaled to the empirical variance of $\vd$.

\minisection{Posterior estimate}
For the selected $(\ell^\star, \sigma_\epsilon^{2,\star})$, the posterior mean of the stacked prior-error vector $\boldsymbol{\epsilon}$ is
\begin{equation}
    \hat{\boldsymbol{\epsilon}}
    \;=\; \hat{\mu}\,\mathbf{1} + \mK_\ell\,\mC^{-1}(\vd - \hat{\mu}\,\mathbf{1}),
    \label{eq:eps_posterior}
\end{equation}
where $\hat{\boldsymbol\epsilon}\in\mathbb{R}^M$ is the universal-kriging posterior mean of $\boldsymbol\epsilon$ given $\vd$.

The marginal posterior variance at $\vx_i$ is
\begin{equation}
    \begin{aligned}
        \mathrm{Var}\!\left[\epsilon(\vx_i) \mid \vd\right]
        &\;=\; [\mK_\ell]_{ii} - (\mK_\ell\mC^{-1}\mK_\ell)_{ii} \\
        &\quad + \frac{\bigl(1 - (\mathbf{1}^\top\mC^{-1}\mK_\ell)_i\bigr)^2}{A},
    \end{aligned}
    \label{eq:eps_var}
\end{equation}
where the third term reflects residual uncertainty about $\mu$ propagated to $\epsilon(\vx_i)$.

The final \tool{FasterRW} temperature estimate is then
\begin{equation}
    \hat{\vT}_\epsilon \;\triangleq\; \tilde{\vT} - \hat{\boldsymbol{\epsilon}},
    \qquad
    \tilde{\vT} \;\triangleq\; [\tilde{T}(\vx_1),\ldots,\tilde{T}(\vx_M)]^\top,
    \label{eq:fasterrw_final}
\end{equation}
where $\hat{\vT}_\epsilon\in\mathbb{R}^M$ is the prior-error-corrected temperature vector at the query points and $\tilde{\vT}$ is the prior temperature vector. Derivations of \Cref{eq:reml,eq:c_hat,eq:eps_posterior,eq:eps_var} are provided in \Cref{app:fusion_derivation}.

The estimator in \Cref{eq:eps_posterior} can be viewed as a statistically regularized correction of the coarse prior field. The observation vector $\vd$ measures the discrepancy between the prior prediction and the random-walk estimate, the drift $\mu$ absorbs the average bias of the coarse FEM solution, and the RBF covariance $\mK_\ell$ encodes the residual spatial coherence in the prior error. After the tail-reuse fusion (\Cref{eq:fused_estimator_implementation}) has produced $\{\hat{T}_q,\mathrm{Var}[\hat{T}_q]\}_{q=1}^M$, the prior-error refinement only solves an $M$-dimensional Gaussian update involving the $M\times M$ covariance matrix $\mC$, whose Cholesky cost is negligible for $M$ in the tens. Therefore, the quantity reported by \tool{FasterRW} is the prior-error-corrected estimate $\hat{\vT}_\epsilon$ in \Cref{eq:fasterrw_final}.

\section{Experiments}
\label{sec:exp}

In this section, we evaluate \tool{FasterRW} on widely used 3D steady-state thermal simulation benchmarks~\cite{Yang2022:ICCAD} and compare against two random-walk baselines.

\subsection{Experimental Settings}
\label{subsec:exp_settings}

For a fair comparison with existing state-of-the-art methods, we closely follow the setups adopted in prior works~\cite{Dong2025:TCAD,Yang2022:ICCAD,luo2025gpu,Wang2026:DATE}. Due to differences in implementation and hardware, some absolute runtimes may vary slightly, but all methods are evaluated under the same conditions.

\begin{figure}[t!]
    \centering
    \includegraphics[width=.8\linewidth]{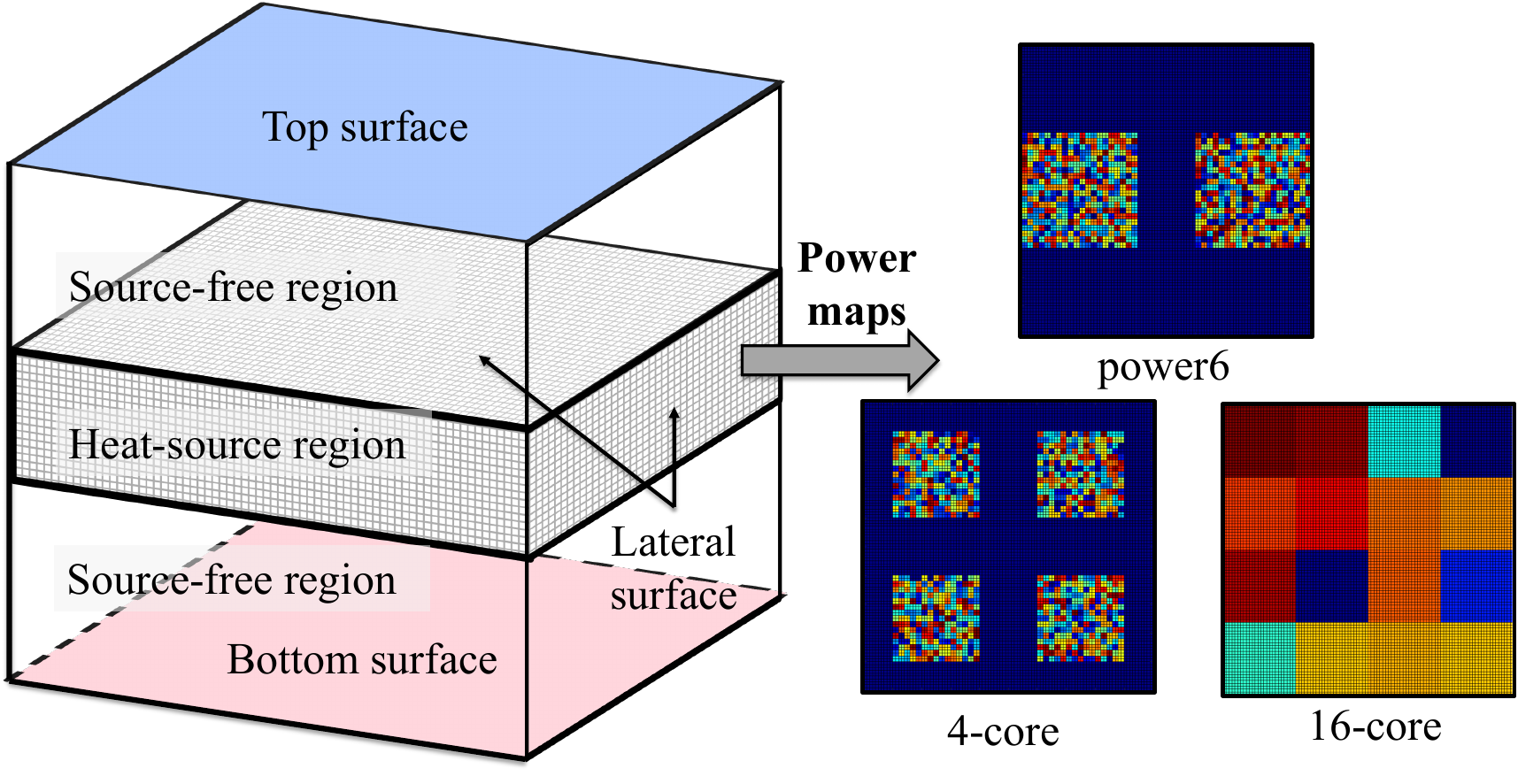}
    \caption{An illustration of test cases and power maps.}
    \label{fig:data}
\end{figure}

\minisection{Test Cases}
Following previous works~\cite{Dong2025:TCAD,Yang2022:ICCAD,luo2025gpu,Wang2026:DATE}, we consider a 3D chip thermal model illustrated in \Cref{fig:data}. The model consists of three layers: the middle layer represents the heat-source region (transistors and metal interconnections), while the top and bottom layers are source-free regions modeling the heat transfer medium and substrate, respectively. The chip measures $2\,\text{cm} \times 2\,\text{cm}$ in the horizontal plane. We study three thickness configurations (top, middle, bottom):
Case~1 $(500,100,500)\,\mu\text{m}$,
Case~2 $(1000,100,1000)\,\mu\text{m}$,
and Case~3 $(500,100,1000)\,\mu\text{m}$.

The heat-source region is discretized into voxels for WOG, with a resolution of $200\,\mu\text{m} \times 200\,\mu\text{m} \times 20\,\mu\text{m}$, consistent with prior work. The thermal conductivities are $395\,\text{W}/(\text{K}\cdot\text{m})$ for source-free regions and $125\,\text{W}/(\text{K}\cdot\text{m})$ for the heat-source region, with an ambient temperature of $293.15\,\text{K}$.

For power maps, we adopt the settings from prior studies, shown in \Cref{fig:data}. We consider three types, \texttt{4-core}, \texttt{16-core}, and \texttt{power6}, with total powers $356\,\text{W}$, $352\,\text{W}$, and $174\,\text{W}$. While the precise power-map values differ slightly from those reported in prior work, leading to minor variations in absolute temperatures, these differences do not affect the relative comparison between algorithms.

Since \tool{FasterRW} focuses on acceleration in cases without Dirichlet boundaries, we follow prior works~\cite{Yang2022:ICCAD,luo2025gpu} and impose Robin boundary conditions on the top and bottom surfaces with convection coefficient $h(\vx) = 8700\,\text{W}/(\text{K}\cdot\text{m}^2)$ for Cases~1 and 2, and $h(\vx) = 4900\,\text{W}/(\text{K}\cdot\text{m}^2)$ for Case~3. Lateral surfaces are modeled as adiabatic Neumann boundaries with zero normal heat flux, $q_n(\vx) = 0$.

\begin{table*}[t!]
    \centering
    \footnotesize
    \caption{Iso-accuracy comparison of \tool{FasterRW} against the PIRW and FastRW baselines. For each (case, $\varepsilon$, method), we report the path count $N$, the mean steps per path (in $10^6$), the mean$\pm$std of the average absolute error over $M=16$ query points and $B=500$ bootstrap trials, the total $N\times\text{steps}$ workload (in $10^9$), and the speedup over PIRW. Speedup is the ratio of $N\times\text{steps}$ workloads; per-stage wallclock costs are reported in \Cref{tab:time}.}
    \label{tab:single}
    \begin{tabular}{c|c|l|cccc|c}
        \toprule
        $\varepsilon$ & Case & Method & $N$ & Steps ($10^6$) & Err$\pm$Std (K) & Workload ($10^9$) & Speedup \\
        \midrule
        \multirow{9}{*}{0.4}
          & \multirow{3}{*}{1}
              & PIRW~\cite{Yang2022:ICCAD}   & 1280 &  5.99 & $0.40\pm0.07$ &  7.66 & $1.0\times$ \\
          &   & FastRW~\cite{Wang2026:DATE}             &  640 &  2.28 & $0.40\pm0.08$ &  1.46 & $5.2\times$ \\
          &   & \tool{FasterRW}                         &  384 &  2.28 & $0.40\pm0.10$ &  0.88 & $\mathbf{8.7\times}$ \\
          \cmidrule(l){2-8}
          & \multirow{3}{*}{2}
              & PIRW~\cite{Yang2022:ICCAD}   & 2560 &  6.00 & $0.40\pm0.07$ & 15.35 & $1.0\times$ \\
          &   & FastRW~\cite{Wang2026:DATE}             & 1280 &  2.29 & $0.40\pm0.07$ &  2.93 & $5.2\times$ \\
          &   & \tool{FasterRW}                         &  640 &  2.29 & $0.38\pm0.08$ &  1.47 & $\mathbf{10.5\times}$ \\
          \cmidrule(l){2-8}
          & \multirow{3}{*}{3}
              & PIRW~\cite{Yang2022:ICCAD}   & 1792 & 10.58 & $0.39\pm0.07$ & 18.96 & $1.0\times$ \\
          &   & FastRW~\cite{Wang2026:DATE}             & 1280 &  4.03 & $0.38\pm0.07$ &  5.16 & $3.7\times$ \\
          &   & \tool{FasterRW}                         &  192 &  4.03 & $0.38\pm0.15$ &  0.77 & $\mathbf{24.5\times}$ \\
        \midrule
        \multirow{9}{*}{0.5}
          & \multirow{3}{*}{1}
              & PIRW~\cite{Yang2022:ICCAD}   &  768 &  5.99 & $0.48\pm0.09$ &  4.60 & $1.0\times$ \\
          &   & FastRW~\cite{Wang2026:DATE}             &  384 &  2.28 & $0.48\pm0.10$ &  0.88 & $5.2\times$ \\
          &   & \tool{FasterRW}                         &  192 &  2.28 & $0.47\pm0.13$ &  0.44 & $\mathbf{10.5\times}$ \\
          \cmidrule(l){2-8}
          & \multirow{3}{*}{2}
              & PIRW~\cite{Yang2022:ICCAD}   & 1280 &  6.00 & $0.48\pm0.09$ &  7.68 & $1.0\times$ \\
          &   & FastRW~\cite{Wang2026:DATE}             &  640 &  2.29 & $0.50\pm0.09$ &  1.47 & $5.2\times$ \\
          &   & \tool{FasterRW}                         &  384 &  2.29 & $0.45\pm0.10$ &  0.88 & $\mathbf{8.7\times}$ \\
          \cmidrule(l){2-8}
          & \multirow{3}{*}{3}
              & PIRW~\cite{Yang2022:ICCAD}   & 1024 & 10.58 & $0.48\pm0.09$ & 10.84 & $1.0\times$ \\
          &   & FastRW~\cite{Wang2026:DATE}             &  640 &  4.03 & $0.48\pm0.09$ &  2.58 & $4.2\times$ \\
          &   & \tool{FasterRW}                         &   96 &  4.03 & $0.49\pm0.22$ &  0.39 & $\mathbf{28.0\times}$ \\
        \bottomrule
    \end{tabular}
\end{table*}

\begin{figure*}[t!]
    \centering
    \subfloat[Case~1]{\includegraphics[width=0.32\linewidth]{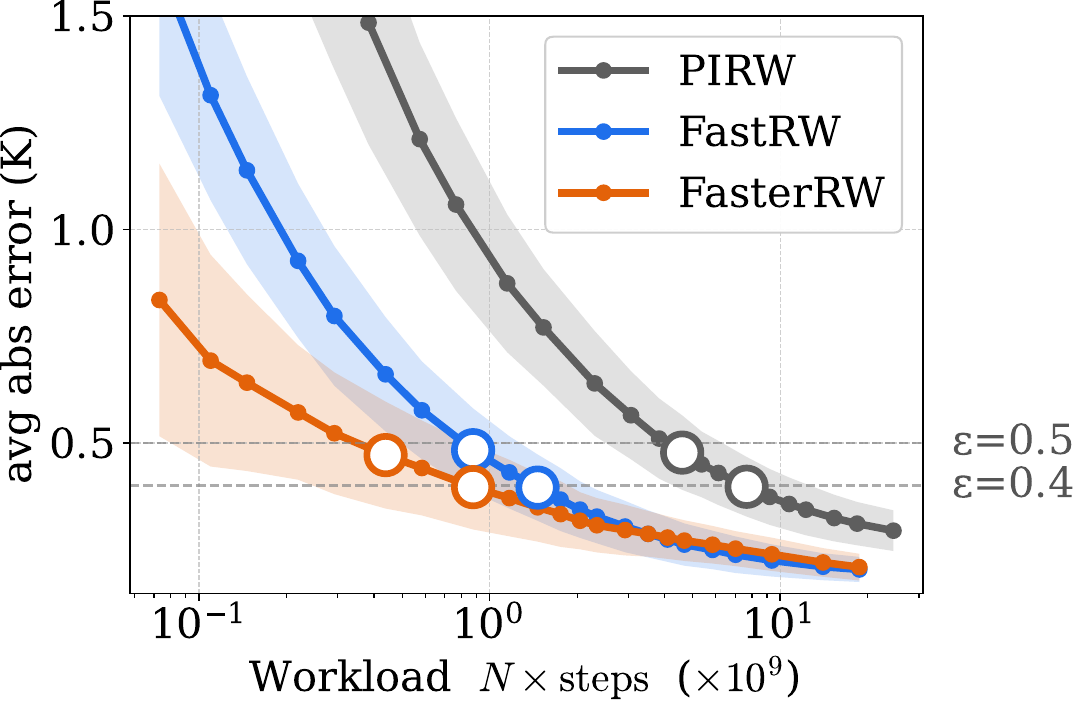}}
    \hfill
    \subfloat[Case~2]{\includegraphics[width=0.32\linewidth]{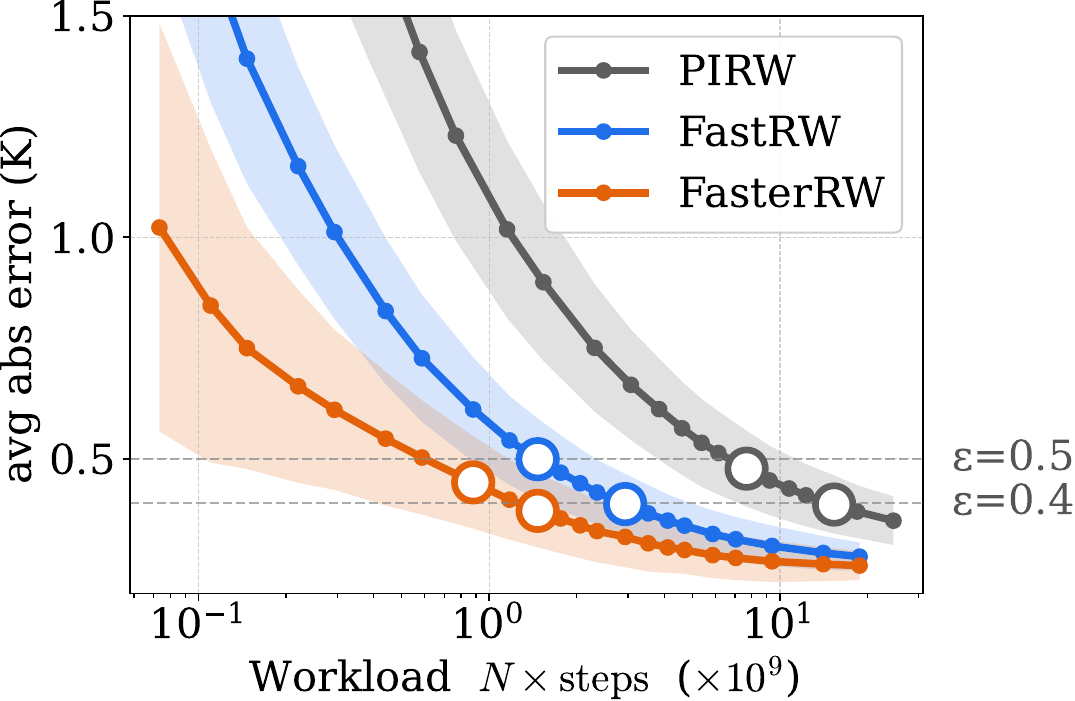}}
    \hfill
    \subfloat[Case~3]{\includegraphics[width=0.32\linewidth]{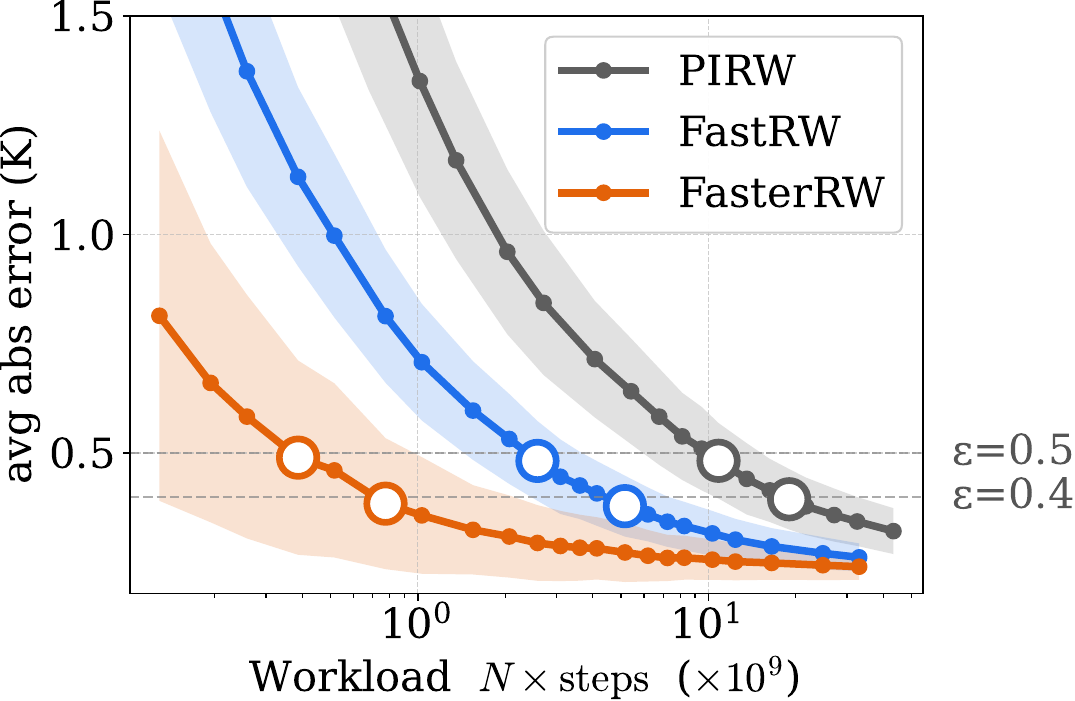}}
    \caption{Bootstrap mean of the average absolute error vs.\ workload. Error bands show $\pm 1$ std over $500$ bootstrap trials. Dashed lines mark the iso-accuracy targets $\varepsilon \in \{0.4, 0.5\}\,\mathrm{K}$; markers indicate the operating points reported in \Cref{tab:single}.}
    \label{fig:bootstrap}
\end{figure*}

\minisection{Implementation Details}
Unless otherwise noted, \tool{FasterRW} uses a truncation threshold $\Lambda = 0.03$. This value follows the threshold-selection rule in \Cref{cor:plugin_bias_uniform}: the worst-case maximum point-wise prior error across the three cases is $5.04\,\mathrm{K}$ (Case~2), so $\Lambda = 0.03$ bounds the threshold-induced correction bias by $0.15\,\mathrm{K}$, well below the iso-accuracy targets $\varepsilon \in \{0.4, 0.5\}\,\mathrm{K}$ used in our experiments. We therefore apply a uniform $\Lambda = 0.03$ across all cases.

For multi-point acceleration, \tool{FasterRW} records at most one tail-reuse pseudo-sample for $T(\vx_j)$ every $1000$ random-walk steps, and only admits pass-through events whose Feynman--Kac weight satisfies $\hat{e}_c(t_k) \ge e_{\min} = 0.3$. In the tail-reuse fusion stage, only cross-target pass-throughs ($j_k \ne i$) are kept, since head and tail samples drawn from the same trajectory would violate the independence assumption of inverse-variance fusion. In the kriging refinement stage, we use observations from all queried locations unless specified otherwise.

The local-time update at Robin and Neumann boundaries follows the boundary-strip allocation used in prior PIRW implementations~\cite{Yang2022:ICCAD,Dong2025:TCAD,luo2025gpu}.

The random walk is implemented in C++ with a Metal compute kernel and runs on the GPU of an Apple M5 Pro workstation. FEM prior estimation, Bayesian tail-reuse fusion, and universal-kriging refinement are CPU stages executed on the same workstation. All baselines use the same C++/Metal codebase and the same hardware, so reported speedups are measured under a single, consistent setup.

To estimate priors for the temperature field, we employ a commercial FEM tool with coarse element discretization. The FEM priors used for the main comparison have approximately $1{,}288$, $2{,}779$, and $2{,}841$ degrees of freedom for Cases~1, 2 and 3. These discretizations are chosen for two reasons: first, they are lightweight enough that the FEM solve time is strictly below $1\,\mathrm{s}$ on each case; second, the resulting prior error is acceptable, with a worst-case point-wise error of $5.04\,\mathrm{K}$ on Case~2. The golden reference is obtained from the finest FEM discretization, whose mesh exceeds $2.5\,\mathrm{M}$ degrees of freedom on every case.

\minisection{Baseline Methods}
We compare \tool{FasterRW} against two random-walk baselines.

\textbf{PIRW}~\cite{Yang2022:ICCAD,luo2025gpu} is the canonical Feynman--Kac solver described in \Cref{sec:pre}; following~\cite{luo2025gpu}, we set its truncation threshold to $\Lambda = 10^{-4}$. PIRW has no prior correction, so its effective tail scale is bounded by the absolute temperature magnitude ($\sim 350\,\mathrm{K}$ in our benchmarks), which explains why it must use a much smaller threshold than \tool{FasterRW} to keep truncation bias small.

\textbf{FastRW}~\cite{Wang2026:DATE} is a recently proposed Feynman--Kac solver that introduces prior-corrected truncation together with Bayesian fusion of cross-point observations. We adopt the same $\Lambda = 0.03$ as in \tool{FasterRW} and the same multi-point hyperparameters, so any remaining gap reflects the kriging refinement contributed by \tool{FasterRW} rather than threshold or fusion choices.

A third method, the hybrid random-walk of~\cite{Liang2014:ASPDAC}, does not reach the sub-kelvin accuracy range targeted in this paper; we discuss this comparison separately in \Cref{subsec:exp_discuss}.

\subsection{Main Results}
\label{subsec:exp_main}

We compare \tool{FasterRW} against the two baselines on all three cases under two iso-accuracy targets, $\varepsilon \in \{0.4, 0.5\}\,\mathrm{K}$. For each $(\text{case}, \varepsilon, \text{method})$ cell, the smallest path count $N$ that meets the target is selected, with per-cell statistics summarized over $M = 16$ uniformly sampled query points and $B = 500$ bootstrap subsamples. The subsamples are drawn with replacement from a single long Monte Carlo run at $N_{\max}$ paths per query, allowing a dense $N$-sweep at modest compute. Results are reported in \Cref{tab:single} and visualized as bootstrap-convergence curves in \Cref{fig:bootstrap}.

\tool{FasterRW} outperforms both baselines on every (case, $\varepsilon$) cell. Against PIRW, \tool{FasterRW} achieves $\mathbf{8.7\text{--}28.0\times}$ compute-unit speedup, with two complementary sources: shorter paths ($\sim 2.6\times$ fewer steps from prior-corrected truncation) and fewer paths (from tail-reuse fusion and kriging refinement together). Against the stronger baseline FastRW~\cite{Wang2026:DATE}, \tool{FasterRW} still cuts the path count by $1.7\text{--}6.7\times$: the kriging step extrapolates the prior-error residual across the query set, an information channel that neither PIRW nor FastRW exploits. The largest gap appears on Case~3 at $\varepsilon = 0.5\,\mathrm{K}$ ($28.0\times$ over PIRW), where the FEM prior is most accurate and the kriging residual is correspondingly smoothest. The bootstrap-convergence curves in \Cref{fig:bootstrap} confirm this ordering: the \tool{FasterRW} curve crosses both iso-accuracy thresholds at strictly smaller $N$ than the FastRW baseline, which in turn crosses at strictly smaller $N$ than PIRW, with the horizontal gap widening at the looser target.

These ratios reflect algorithmic acceleration in random-walk compute units. The corresponding end-to-end wallclock, which additionally charges the FEM-prior and posterior stages, is examined later in this section (see \Cref{tab:time}).

\subsection{Further Discussion}
\label{subsec:exp_discuss}

The headline numbers in \Cref{tab:single} aggregate several distinct sources of acceleration. We now isolate them in turn, first verifying the truncation identity that underpins prior-corrected stopping, then quantifying how the kriging refinement scales with co-batched queries, examining the trade-off between prior accuracy and threshold, stress-testing robustness under deliberately weak priors, reporting the end-to-end wallclock breakdown, and finally positioning \tool{FasterRW} against alternative path-shortening directions.

\begin{figure}[t!]
    \centering
    \subfloat{\includegraphics[height=3.6cm]{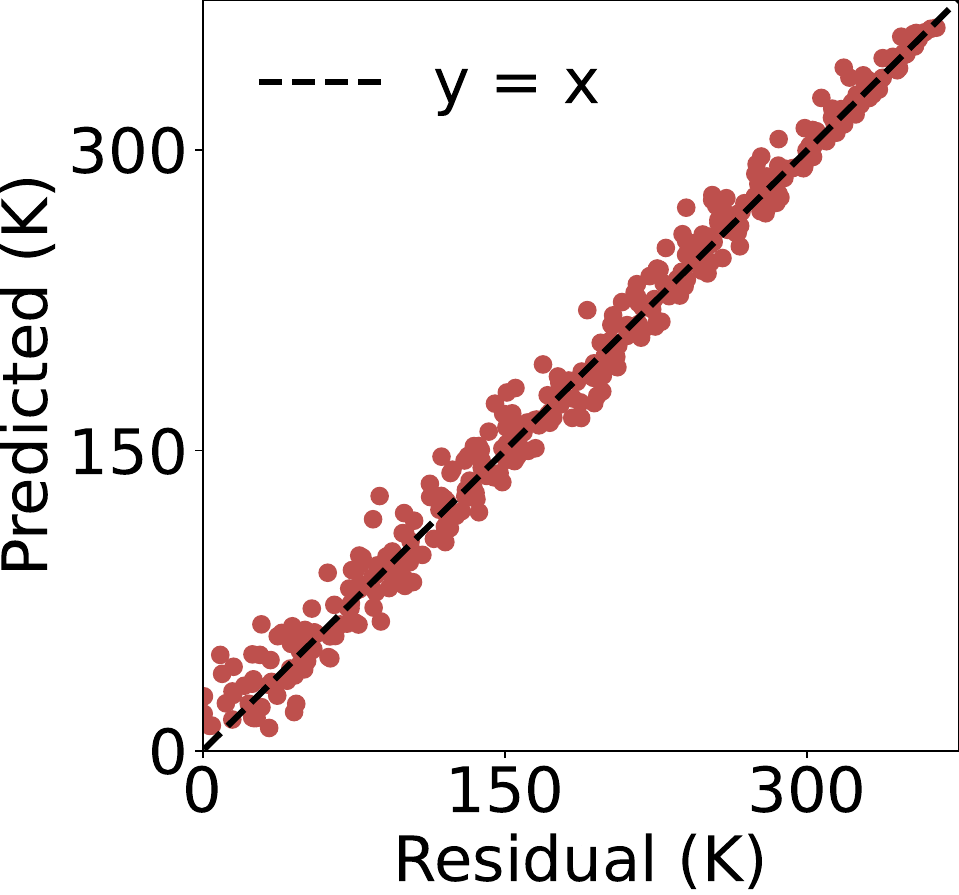}}
    \hfill
    \subfloat{\includegraphics[height=3.6cm]{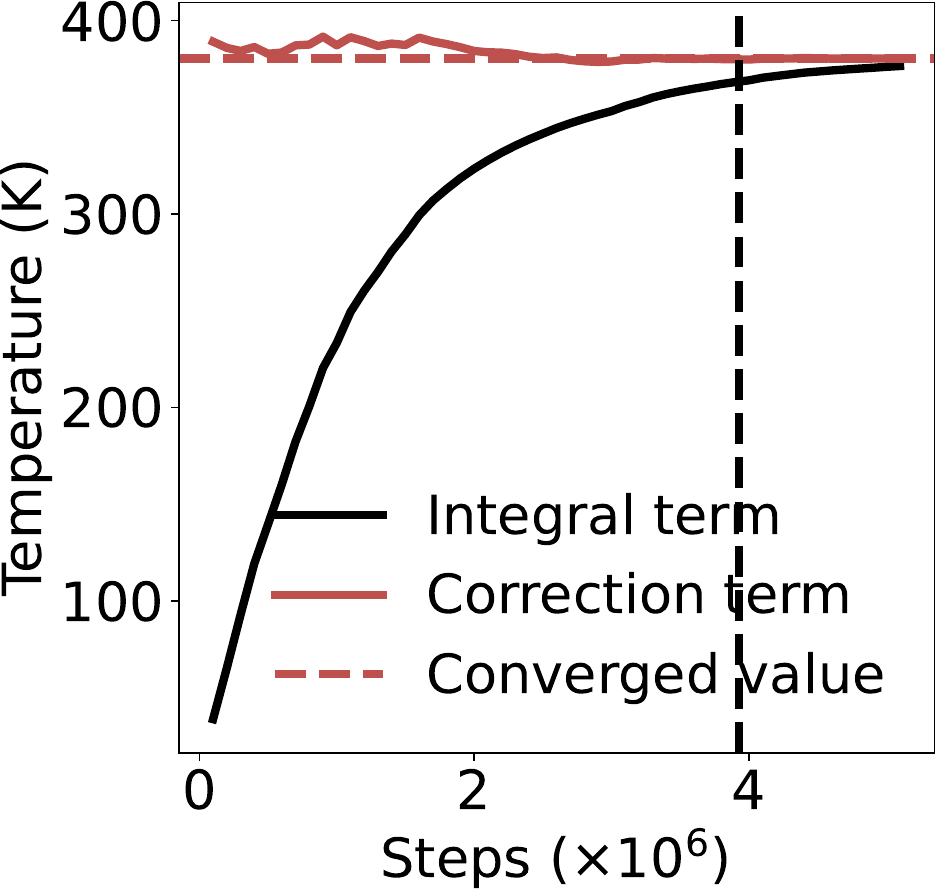}}
    \caption{(Left) Validation of \Cref{corollary:error}: measured residual vs.\ predicted $\hat{e}_c(t)\,T(X_t)$ on sampled paths. (Right) Convergence of $\mathbb{P}_0^{t}$ and the corrected estimate; dashed line marks the truncation step adopted by \tool{FasterRW}.}
    \label{fig:discuss}
\end{figure}

\minisection{Empirical Validation of the Truncation Identity}
Our acceleration rests on \Cref{corollary:error}: the expected truncation residual at time $t$ equals $\hat{e}_c(t)\,T(X_t)$. We first verify this empirically. From the query location in Case~3, we sample $400$ random-walk trajectories and terminate each at a random time so that $\hat{e}_c(t) \in [0.03, 0.95]$. For each trajectory we record both the realized residual and the predicted scaling $\hat{e}_c(t)\,T(X_t)$. The left panel of \Cref{fig:discuss} shows the prediction tracking the realized residuals across the full range of $\hat{e}_c(t)$. The right panel shows the consequence for convergence: the corrected estimate $\mathbb{P}_0^t + \hat{e}_c(t)\,T(X_t)$ stabilizes far earlier than $\mathbb{P}_0^t$ alone. This is the mechanism that lets \tool{FasterRW} terminate paths early without losing accuracy.

\begin{table}[t!]
    \centering
    \footnotesize
    \renewcommand{\arraystretch}{0.95}
    \caption{Per-query equivalent path-count speedup $\hat{N}/N$ as a function of group size on Case~1 ($N=1000$ paths per query, $B=500$ bootstrap trials). Both methods grow monotonically with $G$; \tool{FasterRW} amplifies the cross-query gain through its kriging refinement.}
    \label{tab:multi}
    \begin{tabular}{c|cc|cc}
        \toprule
        \multirow{2}{*}{Group $G$} & \multicolumn{2}{c|}{FastRW~\cite{Wang2026:DATE}} & \multicolumn{2}{c}{\tool{FasterRW}} \\
        & Err. (K) & $\hat{N}/N$ & Err. (K) & $\hat{N}/N$ \\
        \midrule
        1  & $0.407$ & $1.00\times$              & $0.407$ & $1.00\times$ \\
        4  & $0.379$ & $1.20\times \pm 0.11$     & $0.362$ & $1.68\times \pm 0.47$ \\
        8  & $0.357$ & $1.39\times \pm 0.15$     & $0.323$ & $2.26\times \pm 0.71$ \\
        16 & $0.324$ & $1.85\times \pm 0.25$     & $0.307$ & $\mathbf{3.79\times \pm 0.90}$ \\
        \bottomrule
    \end{tabular}\\[2pt]
    % {\scriptsize $^{\dagger}$\,At $G = 1$ the universal-kriging GP has only one training point and fits it exactly, so \tool{FasterRW} degenerates to FastRW.}
\end{table}
\begin{table}[t]
    \centering
    \footnotesize
    \caption{More accurate priors permit larger $\Lambda$ and shorter random walks on Case~1. DoF denotes the degrees of freedom in the FEM prior; $\epsilon_{\max}$ is the global maximum point-wise prior error against the golden reference.}
    \label{tab:tradeoff}
    \begin{tabular}{cc|ccc}
        \toprule
        \multicolumn{2}{c|}{Prior} & \multicolumn{3}{c}{\tool{FasterRW}} \\
        DoF & $\epsilon_{\max}$ (K) & $\Lambda$ & Steps ($10^6$) & RW Time (s) \\
        \midrule
        $699$    & $8.71$ & $0.013$ & $2.83$ & $2.14$ \\
        $1{,}288$  & $3.76$ & $0.030$ & $2.28$ & $1.73$ \\
        $10{,}254$ & $1.26$ & $0.090$ & $\mathbf{1.57}$ & $\mathbf{1.19}$ \\
        \bottomrule
    \end{tabular}
\end{table}

\begin{table*}[t!]
    \centering
    \footnotesize
    \caption{Robustness of \tool{FasterRW} under a rule-of-thumb (uniform-at-ambient) prior on Case~1 at $\varepsilon = 0.4\,\mathrm{K}$. Time/pt is per-query random-walk wallclock on Apple M5 Pro. Speedup is per-query wallclock vs.\ PIRW.}
    \label{tab:weakprior}
    \begin{tabular}{l|lc|cccccc}
        \toprule
        Method & Prior & $\epsilon_{\max}$ (K) & $\Lambda$ & Paths & Steps ($10^6$) & Time/pt (s) & Speedup \\
        \midrule
        PIRW~\cite{Yang2022:ICCAD}      & none           & N/A & $10^{-4}$ & $1280$ & $5.99$ & $5.78$ & $1.00\times$ \\
        FastRW~\cite{Wang2026:DATE}     & rule-of-thumb  & $43.7$     & $10^{-3}$ & $1024$ & $4.49$ & $3.48$ & $1.66\times$ \\
        \tool{FasterRW}                 & rule-of-thumb  & $43.7$     & $10^{-3}$ & $512$  & $4.49$ & $1.74$ & $3.32\times$ \\
        FastRW~\cite{Wang2026:DATE}     & FEM            & $3.76$     & $0.030$   & $640$  & $2.28$ & $1.11$ & $5.21\times$ \\
        \tool{FasterRW}                 & FEM            & $3.76$     & $0.030$   & $384$  & $2.28$ & $0.66$ & $\mathbf{8.76\times}$ \\
        \bottomrule
    \end{tabular}
\end{table*}

\minisection{Effect of Group Size}
The kriging refinement in \tool{FasterRW} extrapolates the prior-error residual across co-batched queries; its benefit therefore depends on how many queries are batched together. The FastRW baseline only fuses tail-reuse pseudo-samples, without this extrapolation step. We quantify the resulting gap by holding the path count fixed at $N = 1000$ per query on Case~1 and partitioning the $M = 16$ queries into disjoint subgroups of size $G \in \{1, 4, 8, 16\}$. Each pseudo-sample is admitted only if both its source and target lie in the same subgroup, matching the user-facing scenario of co-querying $G$ points. We then report the per-query equivalent path count $\hat{N}/N = \mathrm{Var}_{\text{single}}/\mathrm{Var}_{\text{fused}}$ averaged over queries; this ratio follows directly from the law-of-large-numbers scaling $\mathrm{Var}[\bar T] = \sigma^2/N$ in \Cref{eq:largenumber}, so a $k\times$ variance reduction is equivalent to running $k\times$ more independent paths. By construction $\hat{N}/N = 1$ at $G = 1$. The results are summarized in \Cref{tab:multi}.

The FastRW baseline grows from $1.00\times$ to $1.85\times$ as $G$ goes from $1$ to $16$: the number of cross-target tail-reuse pass-throughs scales roughly with $G - 1$, but each pass-through carries limited per-path information under $e_{\min} = 0.3$. \tool{FasterRW} grows from $1.00\times$ to $\mathbf{3.79\times}$, roughly doubling the FastRW gain at $G = 16$. The reason is structural: the kriging step exploits the spatial smoothness of the prior-error residual, a signal that scales with the number of training points rather than with per-path pass-through statistics. The mean absolute error decreases in parallel for both methods, consistent with the variance reduction.

\minisection{Prior Accuracy versus Truncation Threshold}
By \Cref{eq:single}, a smaller prior error $\epsilon_{\max}$ permits a larger truncation threshold $\Lambda$, which directly shortens paths. The threshold-selection rule \Cref{cor:plugin_bias_uniform} ties this trade-off to a worst-case bias bound. We test it on Case~1 by varying the FEM prior DoF. We pick three DoFs whose worst-case point-wise prior error $\epsilon_{\max}$ drops roughly monotonically (in general, FEM prior error depends on mesh layout and is not strictly monotone in DoF). For each row, $\Lambda$ is chosen so that $\Lambda \times \epsilon_{\max}$ stays near $0.11\,\mathrm{K}$, well below both iso-accuracy targets. \Cref{tab:tradeoff} summarizes the result.

Improving prior accuracy nearly halves both the mean steps per path (from $2.83 \times 10^6$ to $1.57 \times 10^6$) and the per-query random-walk wallclock (from $2.14\,\mathrm{s}$ to $1.19\,\mathrm{s}$). Moreover, the FEM prior is computed once per case and can be reused across repeated queries, so in any workflow that probes the same chip many times, paying for a finer prior is essentially a free lunch.

\minisection{Robustness to Weak Priors}
The previous paragraph shows that better priors help. Does \tool{FasterRW} also \emph{need} them? We stress-test this on Case~1 with a deliberately crude rule-of-thumb prior: a uniform temperature equal to ambient across all voxels. This prior has maximum error $43.7\,\mathrm{K}$, since the hottest voxel sits $\sim 44\,\mathrm{K}$ above ambient. Under \Cref{cor:plugin_bias_uniform}, $\Lambda = 0.001$ bounds the threshold-induced correction bias by $0.001 \times 43.7 \approx 0.044\,\mathrm{K}$, well below the $\varepsilon = 0.4\,\mathrm{K}$ target, while remaining $10\times$ larger than PIRW's threshold. \Cref{tab:weakprior} reports \tool{FasterRW} and the FastRW baseline under this rule-of-thumb prior, alongside the FEM-prior reference.

Even under this crude $43.7\,\mathrm{K}$-error prior, \tool{FasterRW} reaches a $3.32\times$ wallclock speedup over PIRW---and the FastRW baseline still gets $1.66\times$. This confirms the key implication of \Cref{eq:single}: the prior need not be accurate enough to solve the problem on its own. It only needs to shrink the residual scale from the absolute temperature magnitude ($\sim 350\,\mathrm{K}$) to the prior-error magnitude; the random-walk correction recovers the remaining accuracy. Switching to the FEM prior shrinks the residual scale by another order of magnitude and unlocks roughly $3\times$ more for both methods: $\mathbf{8.76\times}$ for \tool{FasterRW} and $5.21\times$ for FastRW. \tool{FasterRW}'s $\sim 2\times$ advantage over FastRW persists across the prior-quality spectrum, because the kriging step exploits residual smoothness whether the residual is small (FEM prior) or large (the rule-of-thumb prior leaves the entire physics field as residual, still spatially smooth, just full-amplitude).

\minisection{End-to-end Wallclock Breakdown}
\tool{FasterRW} adds three stages over PIRW: a FEM-prior stage, a tail-reuse fusion stage, and a kriging refinement stage. The FastRW baseline shares the first two and lacks only the kriging refinement. We report per-query wallclock on Case~1 at $\varepsilon = 0.4\,\mathrm{K}$, broken down by stage, in \Cref{tab:time}.

\begin{table}[t!]
    \centering
    \footnotesize
    \caption{Per-query wallclock breakdown on Case~1 at $\varepsilon = 0.4\,\mathrm{K}$, measured on Apple M5 Pro. The FEM prior is amortized over $M = 16$ queries and reported as the steady-state MUMPS linear-solve cost\textsuperscript{$\dagger$} (excluding one-time mesh-build overhead).}
    \label{tab:time}
    \begin{tabular}{l|ccc}
        \toprule
        Stage (s) & PIRW & FastRW & \tool{FasterRW} \\
        \midrule
        FEM prior          & ---   & $<0.06$\textsuperscript{$\dagger$} & $<0.06$\textsuperscript{$\dagger$} \\
        Random walk        & $5.78$ & $1.11$ & $0.66$ \\
        Tail-reuse fusion  & ---   & $0.02$ & $0.01$ \\
        Kriging refinement & ---   & ---    & $0.02$ \\
        \midrule
        Total per query    & $\mathbf{5.78}$ & $<1.19$ & $\mathbf{<0.76}$ \\
        Wallclock speedup  & $1.00\times$ & $>4.88\times$ & $\mathbf{>7.61\times}$ \\
        \bottomrule
    \end{tabular}\\[2pt]
    {\scriptsize $^{\dagger}$\,Commercial tool logs solve time as $0\,\mathrm{s}$ (integer-second rounding) for the DoF$=1288$ prior; the total is therefore $<1\,\mathrm{s}$, i.e., $<0.06\,\mathrm{s}$ amortized over $M=16$ queries.}
\end{table}

End-to-end, \tool{FasterRW} reaches $\mathbf{>7.61\times}$ wallclock speedup over PIRW on this configuration. This is slightly smaller than the $8.7\times$ compute-unit speedup reported in \Cref{tab:single}; the residual gap is the FEM-prior cost, charged here as the steady-state MUMPS linear solve ($<0.06\,\mathrm{s}$ per query, amortized over $M = 16$ queries). The tail-reuse fusion and kriging refinement together contribute $0.03\,\mathrm{s}$ per query, confirming that the speedup gains arise from path- and step-count reductions, not from a cheaper backend. For context, a ground-truth FEM solve on Case~1 takes more than $450\,\mathrm{s}$, two orders of magnitude longer than the entire \tool{FasterRW} per-query budget, which confirms that the coarse FEM serves as an inexpensive prior rather than a replacement for the random-walk solver.

\minisection{Comparison with Alternative Methods}
We briefly position \tool{FasterRW} against two alternative path-shortening directions that we considered but excluded from the main tables.

The hybrid random-walk method of~\cite{Liang2014:ASPDAC} replaces parts of the Feynman--Kac formulation with a coarser heuristic for Robin boundaries. We re-implemented it in our setting; the maximum error reaches $7.39\,\mathrm{K}$ on Case~1, well outside the sub-kelvin range targeted in this paper, and the other cases show similar behavior. We therefore retain PIRW as the strongest direct baseline in the same reflected Feynman--Kac regime.

A second direction is the empirical curve fit used by the GPU PIRW variant~\cite{luo2025gpu}:
\begin{subequations}
\begin{align}
\mathbb{P}_0^t &= T(\vx) - a\,\exp(b\,t + c), \label{eq:iseda1}\\
\hat{T}(\vx) &= \frac{\mathbb{P}_0^{t-l}\,\mathbb{P}_0^{t-l} - \mathbb{P}_0^t\,\mathbb{P}_0^{t-2l}}
{2\,\mathbb{P}_0^{t-l} - \mathbb{P}_0^t - \mathbb{P}_0^{t-2l}}, \label{eq:iseda2}
\end{align}
\end{subequations}
where $l$ is a fixed sampling gap and $a$, $b$, $c$ are unknown constants. Comparing \Cref{eq:iseda1} with \Cref{eq:twopart,eq:error} shows that \Cref{eq:iseda1} is an approximation that assumes a strict map between the functional $\hat{e}_c(t)$ and the step count $t$. In practice, random fluctuations along walks make \Cref{eq:iseda2} unstable, and we did not obtain reasonable results in our implementations. \Cref{eq:iseda1,eq:iseda2} also offer no analogue of \tool{FasterRW}'s multi-point fusion or threshold-selection rule, and so we exclude this approach from the main tables.

\section{Conclusion}
This paper introduced \tool{FasterRW}, a random-walk solver that accelerates steady-state thermal analysis by quantitatively analyzing truncation error and incorporating both prior estimation and posterior correction. The truncation residual identity explains why prior-assisted truncation can use substantially larger thresholds while keeping the correction error controlled by the prior error rather than the absolute temperature scale, leading to an explicit threshold-selection rule. The multi-point formulation further reuses the tail segment of every random-walk path as an approximately unbiased pseudo-sample for the other query points and, in prior-error coordinates, refines the result through a universal-kriging Gaussian-process prior on the prior-error field. By leveraging coarse FEM-based priors together with tail-reuse fusion and kriging refinement, \tool{FasterRW} achieves $8.7$--$28\times$ compute-unit speedup and over $7.6\times$ end-to-end wallclock speedup over PIRW across the studied cases while maintaining sub-kelvin accuracy, with negligible posterior-correction overhead. Expanded experiments validate the residual behavior, path convergence, group-size effect, prior-accuracy trade-off, robustness under weak priors, and end-to-end runtime breakdown. These results highlight \tool{FasterRW} as a practical tool for efficient hotspot estimation, and future extensions may explore more accurate priors, including learning-based priors, transient conduction, and hardware acceleration.

\section*{Acknowledgment}
The authors used ChatGPT and Codex only for language polishing and editorial assistance, including minor wording suggestions. All technical content and results were created and approved by the authors.

\bibliographystyle{IEEEtran}
\bibliography{ref/Top-sim, ref/bib}

\appendices
\section{Proof of the Truncation Residual Identity}
\label{app:truncation_proof}

This appendix proves \Cref{thm:truncation_residual}. The same argument applies to the discrete-time Markov chain estimators used in implementation: conditioning at a stopping time and restarting from the truncation state yields the residual identity, with the continuous reflected diffusion serving as a limiting interpretation.

\subsection{Proof}
Let $\{\mathcal{F}_t\}_{t\ge 0}$ be the natural filtration generated by the reflected process $\{X_t\}_{t\ge 0}$ and its local time $\{L(t)\}_{t\ge 0}$. Consider a stopping time $\tau$ such that $\tau<t_D$ on the event of interest, and define the post-$\tau$ shifted process $\tilde{X}_s \triangleq X_{\tau+s}$ for $s\ge 0$. Let $\tilde{L}(s)$ denote the boundary local time accumulated by $\tilde{X}$ after time $\tau$, so that $L(\tau+s)=L(\tau)+\tilde{L}(s)$. Define the shifted weight
\begin{equation}
    \tilde{e}_c(s) \triangleq \exp\!\left(-\int_0^{s} c(\tilde{X}_u)\,\mathrm{d}\tilde{L}(u)\right).
\end{equation}
By construction,
\begin{equation}
    \hat{e}_c(\tau+s)
    =
    \hat{e}_c(\tau)\,\tilde{e}_c(s).
    \label{eq:weight_shift}
\end{equation}

Define $\tilde{\mathbb{P}}_0^\infty$ as the full Feynman--Kac functional evaluated along the shifted path $\tilde{X}$ starting from $\tilde{X}_0=X_\tau$, with the shifted weight $\tilde{e}_c(\cdot)$. By \Cref{eq:weight_shift}, the truncation tail along the original path factors as
\begin{equation}
    \mathbb{P}_\tau^\infty = \hat{e}_c(\tau)\,\tilde{\mathbb{P}}_0^\infty.
    \label{eq:tail_factor}
\end{equation}
Condition on $\mathcal{F}_\tau$. By the strong Markov property of the reflected process at the stopping time $\tau$, the conditional law of the shifted process $(\tilde{X}_s,\tilde{L}(s))_{s\ge 0}$ given $\mathcal{F}_\tau$ is the same as that of a fresh reflected process started from $X_\tau$, and is independent of the past beyond the dependence through $X_\tau$. Therefore, by the definition of $T(\cdot)$ in \Cref{eq:FK-formula},
\begin{equation}
    \mathbb{E}\!\left[\tilde{\mathbb{P}}_0^\infty \mid \mathcal{F}_\tau\right] = T(X_\tau).
    \label{eq:shifted_fk_expectation}
\end{equation}
Combining \Cref{eq:tail_factor,eq:shifted_fk_expectation} yields
\begin{equation}
    \mathbb{E}\!\left[\mathbb{P}_{\tau}^{\infty}\mid \mathcal{F}_{\tau}\right]
    =
    \hat{e}_c(\tau)\,T(X_\tau),
\end{equation}
which completes the proof.

\section{Bias--Variance Analysis of Prior-Based Truncation Correction}
\label{app:plugin_bias_variance}

This appendix extends the prior-correction bias result in \Cref{thm:plugin_bias,cor:plugin_bias_uniform} with a mean-squared-error (MSE) decomposition, an explicit bias--variance split after averaging, and a discussion of how the truncation threshold $\Lambda$ trades the two. The conclusion is that, in the operating regime targeted by \tool{FasterRW}, the per-path variance $\sigma^2(\vx)$ has only a weak dependence on $\Lambda$, so $\Lambda$ can be chosen from the bias bound and $N$ from the variance budget without coupling the two knobs.

\subsection{MSE Decomposition}
The prior-corrected estimator in \Cref{eq:single} can be viewed as the ideal unbiased estimator in \Cref{eq:ideal_estimator} plus a weighted prior-error term. Let $\Delta(\vx) \triangleq \hat{T}_i^\star(\vx) - T(\vx)$ denote the intrinsic Monte Carlo error of the ideal estimator. From \Cref{eq:plugin_error_pathwise},
\begin{equation}
    \hat{T}_i(\vx) - T(\vx) = \Delta(\vx) + \hat{e}_c(\tau)\,\epsilon(X_\tau).
\end{equation}
Therefore, the per-path MSE admits the decomposition
\begin{align}
    &\mathbb{E}\!\left[(\hat{T}_i(\vx) - T(\vx))^2\right] \notag\\
    &=
    \underbrace{\mathbb{E}\!\left[\Delta(\vx)^2\right]}_{\text{ideal MC variance}}
    + 2\,\underbrace{\mathbb{E}\!\left[\Delta(\vx)\,\hat{e}_c(\tau)\,\epsilon(X_\tau)\right]}_{\text{cross term}} \notag\\
    &\quad
    + \underbrace{\mathbb{E}\!\left[\hat{e}_c(\tau)^2\,\epsilon(X_\tau)^2\right]}_{\text{plug-in contribution}}.
    \label{eq:plugin_mse_decomp}
\end{align}
Using Cauchy--Schwarz, the cross term is bounded by
\begin{align}
    &\left|\mathbb{E}\!\left[\Delta(\vx)\,\hat{e}_c(\tau)\,\epsilon(X_\tau)\right]\right| \notag\\
    &\quad
    \le
    \sqrt{\mathbb{E}\!\left[\Delta(\vx)^2\right]}\,
    \sqrt{\mathbb{E}\!\left[\hat{e}_c(\tau)^2\,\epsilon(X_\tau)^2\right]},
\end{align}
which yields the convenient upper bound
\begin{align}
    &\mathbb{E}\!\left[(\hat{T}_i(\vx) - T(\vx))^2\right] \notag\\
    &\quad
    \le
    \left(
        \sqrt{\mathbb{E}\!\left[\Delta(\vx)^2\right]}
        +
        \sqrt{\mathbb{E}\!\left[\hat{e}_c(\tau)^2\,\epsilon(X_\tau)^2\right]}
    \right)^2.
    \label{eq:plugin_mse_bound}
\end{align}
This bound is tight only when $\Delta(\vx)$ and $\hat{e}_c(\tau)\,\epsilon(X_\tau)$ are perfectly correlated. This is not expected in our setting: $\Delta(\vx)$ aggregates Monte Carlo noise over the entire reflected Brownian path including the source/boundary excursions, while $\hat{e}_c(\tau)\,\epsilon(X_\tau)$ depends only on the prior-error field evaluated at the truncation location. Treating the two contributions as approximately uncorrelated gives the working approximation
\begin{equation}
    \mathbb{E}\!\left[(\hat{T}_i(\vx) - T(\vx))^2\right]
    \approx
    \mathbb{E}\!\left[\Delta(\vx)^2\right]
    + \mathbb{E}\!\left[\hat{e}_c(\tau)^2\,\epsilon(X_\tau)^2\right],
    \label{eq:plugin_mse_approx}
\end{equation}
which separates the contribution of the underlying Monte Carlo estimator from the contribution of the prior-error plug-in.

\subsection{Explicit Bias--Variance Split After Averaging}
\tool{FasterRW} averages $N$ independent paths as in \Cref{eq:avg}. Let $\bar{T}(\vx) \triangleq \frac{1}{N}\sum_{i=1}^N \hat{T}_i(\vx)$. The standard bias--variance decomposition gives
\begin{equation}
    \mathbb{E}\!\left[(\bar{T}(\vx) - T(\vx))^2\right]
    =
    \underbrace{\bigl(\mathbb{E}\!\left[\hat{e}_c(\tau)\,\epsilon(X_\tau)\right]\bigr)^2}_{\text{squared bias}}
    +
    \underbrace{\frac{\sigma^2(\vx)}{N}}_{\text{variance}},
    \label{eq:bv_split}
\end{equation}
where the bias term is taken from \Cref{eq:plugin_bias} and the variance from \Cref{eq:largenumber}. Under the threshold rule $\hat{e}_c(\tau)\le\Lambda$, the squared-bias factor is uniformly bounded by $\Lambda^2\,\epsilon_{\max}^2$ as in \Cref{cor:plugin_bias_uniform}. The variance factor decays as $1/N$ and is governed by the per-path variance $\sigma^2(\vx)$, which is itself a function of $\Lambda$ analyzed below.

\subsection{Behavior of $\sigma^2(\vx)$ Under $\Lambda$}
The per-path variance $\sigma^2(\vx) = \mathrm{Var}[\hat{T}_i(\vx)]$ depends on $\Lambda$ through the stopping location and the returned continuation value:
\begin{itemize}
    \item \emph{Shorter pre-truncation paths.} Increasing $\Lambda$ triggers the condition $\hat{e}_c(\tau)\le\Lambda$ earlier, so the partial Feynman--Kac sum $\mathbb{P}_0^\tau$ accumulates fewer source and boundary contributions.
    \item \emph{Larger continuation value.} Earlier stopping also increases the magnitude of the returned continuation term $\hat{e}_c(\tau)\,\tilde{T}(X_\tau)$, whose realization depends on the random terminal location $X_\tau$.
\end{itemize}
The continuation term itself is on the temperature scale; it is not bounded by $\Lambda\epsilon_{\max}$. What the prior-error analysis controls is the extra discrepancy between the practical estimator and the ideal continuation estimator, namely $\hat{e}_c(\tau)\epsilon(X_\tau)$ in \Cref{eq:plugin_error_pathwise}. Under $\hat{e}_c(\tau)\le\Lambda$ this additional error is bounded by $\Lambda\epsilon_{\max}$, while the remaining variance of the ideal continuation estimator is governed by the reflected process and the geometry. Empirically, $\sigma^2(\vx)$ exhibits only a weak dependence on $\Lambda$ in the tested range; this is what allows the path-count knob $N$ to be chosen essentially independently of $\Lambda$.

\subsection{Practical Implications}
The decoupling above justifies the two-step tuning protocol used in \tool{FasterRW}:
\begin{enumerate}
    \item Given a target bias tolerance $b_{\mathrm{tol}}$ and a prior error bound $\epsilon_{\max}$, choose $\Lambda \le b_{\mathrm{tol}}/\epsilon_{\max}$ via \Cref{cor:plugin_bias_uniform}.
    \item Given a target variance tolerance, choose $N$ so that $\sigma^2(\vx)/N$ is within budget via \Cref{eq:largenumber}.
\end{enumerate}
We do not attempt to solve for the MSE-optimal $\Lambda$ jointly with $N$ in closed form: the per-path variance $\sigma^2(\vx)$ is not available analytically for the reflected Brownian process under mixed boundary conditions, and the dependence of both path length and terminal location on $\Lambda$ is geometry-specific. Instead, the bias bound provides a conservative, deployment-ready rule, and the variance is handled separately through path averaging. The weak-prior experiment in \Cref{tab:weakprior} validates this protocol even when $\epsilon_{\max}$ is intentionally large.

\section{REML Derivation for Universal Kriging in Prior-Error Coordinates}
\label{app:fusion_derivation}

This appendix derives the closed-form expressions for the REML marginal likelihood and the universal-kriging posterior used in \Cref{subsec:prior_error_coordinates}. We adopt the model in \Cref{eq:prior_error_obs,eq:uk_prior}:
\begin{equation}
\begin{aligned}
    \vd \mid \boldsymbol{\epsilon}
    &\sim\mathcal{N}(\boldsymbol{\epsilon},\,\mSigma_{\mathrm{MC}}),\\
    \boldsymbol{\epsilon}
    &= \mu\,\mathbf{1} + \mathbf{g},\\
    \mathbf{g}
    &\sim\mathcal{N}(\mathbf{0},\,\mK_\ell),
    \qquad p(\mu)\propto 1,
\end{aligned}
\end{equation}
where $\vd\in\mathbb{R}^M$ is the observation vector, $\boldsymbol{\epsilon}$ the stacked prior-error vector, $\mSigma_{\mathrm{MC}}$ the diagonal Monte Carlo noise covariance, $\mu$ the unknown global drift, and $\mathbf{g}$ the zero-mean spatial residual with RBF covariance $\mK_\ell$. Define the total covariance $\mC \triangleq \mK_\ell + \mSigma_{\mathrm{MC}}$.

\subsection{REML Marginal Likelihood}
Marginalising over $\boldsymbol{\epsilon}$ gives the conditional distribution $\vd\mid \mu \sim \mathcal{N}(\mu\,\mathbf{1},\,\mC)$. The joint density of $(\vd, \mu)$ is, up to a constant,
\begin{equation}
    p(\vd,\mu) \;\propto\; (\det\mC)^{-1/2}\exp\!\left(-\tfrac{1}{2}(\vd-\mu\mathbf{1})^\top\mC^{-1}(\vd-\mu\mathbf{1})\right).
\end{equation}
Expanding the quadratic in $\mu$ and defining $A \triangleq \mathbf{1}^\top\mC^{-1}\mathbf{1}$ and $B \triangleq \mathbf{1}^\top\mC^{-1}\vd$ yields
\begin{equation}
    (\vd-\mu\mathbf{1})^\top\mC^{-1}(\vd-\mu\mathbf{1})
    = A\!\left(\mu - \tfrac{B}{A}\right)^{2} + \vd^\top\mC^{-1}\vd - \tfrac{B^2}{A}.
\end{equation}
Integrating $\mu$ over $\mathbb{R}$ against the flat prior,
\begin{equation}
    \int_{\mathbb{R}}\!\exp\!\left(-\tfrac{1}{2}A(\mu-B/A)^2\right)\,\mathrm{d}\mu = \sqrt{2\pi/A},
\end{equation}
gives the marginal density of $\vd$:
\begin{align}
    p(\vd) &\propto (\det\mC)^{-1/2}\,A^{-1/2}
        \exp\!\left(-\tfrac{1}{2}\,\vd^\top\mC^{-1}\vd + \tfrac{1}{2}\tfrac{B^2}{A}\right) \notag\\
    &= (\det\mC)^{-1/2}\,A^{-1/2}
        \exp\!\left(-\tfrac{1}{2}\mathbf{r}^\top\mC^{-1}\mathbf{r}\right),
\end{align}
where $\mathbf{r}\triangleq\vd-\hat{\mu}\mathbf{1}$ and $\hat{\mu} \triangleq B/A$ is the generalized-least-squares estimate of the drift. Taking the logarithm reproduces \Cref{eq:reml,eq:c_hat}. The estimator $\hat{\mu}$ is the best linear unbiased estimator (BLUE) of $\mu$ under the known $\mC$.

\subsection{Posterior Mean and Variance of $\boldsymbol{\epsilon}$}
By Bayes' rule, conditional on $\mu$, the posterior of $\boldsymbol{\epsilon}$ given $\vd$ is jointly Gaussian. From the linear-Gaussian observation model
\begin{equation}
    \vd\mid\boldsymbol{\epsilon}\sim\mathcal{N}(\boldsymbol{\epsilon},\,\mSigma_{\mathrm{MC}}),
    \qquad
    \boldsymbol{\epsilon}\mid \mu \sim \mathcal{N}(\mu\,\mathbf{1},\,\mK_\ell),
\end{equation}
standard conjugacy gives
\begin{align}
    \mathbb{E}[\boldsymbol{\epsilon} \mid \vd, \mu]
    &= \mu\,\mathbf{1} + \mK_\ell\,\mC^{-1}(\vd - \mu\,\mathbf{1}),
    \label{eq:cond_post_mean}\\
    \mathrm{Cov}[\boldsymbol{\epsilon}\mid\vd, \mu]
    &= \mK_\ell - \mK_\ell\,\mC^{-1}\,\mK_\ell.
    \label{eq:cond_post_cov}
\end{align}
The posterior of $\mu\mid\vd$ is Gaussian with mean $\hat{\mu}$ and variance $1/A$. Marginalising over $\mu$, the posterior mean of the stacked prior-error vector is
\begin{equation}
    \mathbb{E}[\boldsymbol{\epsilon}\mid\vd]
    = \hat{\mu}\,\mathbf{1} + \mK_\ell\,\mC^{-1}(\vd - \hat{\mu}\,\mathbf{1}),
\end{equation}
which is \Cref{eq:eps_posterior}.

For the posterior variance, write $\epsilon_i \triangleq \epsilon(\vx_i)$ for the $i$-th component of $\boldsymbol{\epsilon}$. The law of total variance gives, at each index $i$,
\begin{align}
    \mathrm{Var}[\epsilon_i\mid\vd]
    &= \mathbb{E}_\mu\!\left[\mathrm{Var}[\epsilon_i\mid\vd, \mu]\right] + \mathrm{Var}_\mu\!\left[\mathbb{E}[\epsilon_i\mid\vd, \mu]\right] \notag\\
    &= \bigl(\mK_\ell - \mK_\ell\mC^{-1}\mK_\ell\bigr)_{ii} \notag\\
    &\quad + \bigl(1 - (\mathbf{1}^\top\mC^{-1}\mK_\ell)_i\bigr)^2 \,\mathrm{Var}[\mu\mid\vd].
\end{align}
Substituting $\mathrm{Var}[\mu\mid\vd] = 1/A$ gives \Cref{eq:eps_var}. The second term captures the residual uncertainty about the global drift $\mu$ propagated to $\epsilon_i$; it vanishes precisely when $(\mathbf{1}^\top\mC^{-1}\mK_\ell)_i = 1$, i.e., when the row of $\mC^{-1}\mK_\ell$ sums to one.

\subsection{Algorithmic Use}
In practice the matrix inversions in \Cref{eq:reml,eq:eps_posterior,eq:eps_var} are computed once per hyperparameter candidate via a single Cholesky factorisation of $\mC$. The selection over a grid in $(\ell,\sigma_\epsilon^2)$ then costs $O(|grid|\cdot M^3)$, which is negligible for $M$ in the tens.

\end{document}